\documentclass[11pt]{article}

\usepackage[final]{acl}

\usepackage{times}
\usepackage{latexsym}
\usepackage[T1]{fontenc}
\usepackage[utf8]{inputenc}
\usepackage{microtype}
\usepackage{inconsolata}
\usepackage{enumitem}
\usepackage{amsmath}
\usepackage{tabularray}
\usepackage{graphicx}
\usepackage{booktabs}
\usepackage{multirow}
\usepackage{color}
\usepackage{colortbl}
\usepackage{pifont}
\usepackage{xcolor}
\usepackage{subcaption}
\usepackage{arydshln}
\usepackage{url}

\DeclareMathOperator*{\argmax}{arg\,max}

\title{Bridging the Modality Gap by Similarity Standardization\\with Pseudo-Positive Samples}

\author{
  Shuhei Yamashita \quad Daiki Shirafuji \quad Tatsuhiko Saito \\
  Mitsubishi Electric Corporation \\
  \texttt{\{Yamashita.Shuhei@bc, Shirafuji.Daiki@ay, Saito.Tatsuhiko@db\}} \\
  \texttt{.MitsubishiElectric.co.jp}
}

\begin{document}

\maketitle
\begin{abstract}
  Advances in vision-language models~(VLMs) have enabled effective cross-modality retrieval.  
However, when both text and images exist in the database, similarity scores would differ in scale by modality.
This phenomenon, known as the modality gap, hinders accurate retrieval.
Most existing studies address this issue with manually labeled data, e.g., by fine-tuning VLMs on them.
In this work, we propose a similarity standardization approach with pseudo data construction.
We first compute the mean and variance of the similarity scores between each query and its paired data in text or image modality.
Using these modality-specific statistics, we standardize all similarity scores to compare on a common scale across modalities.
These statistics are calculated from pseudo pairs, which are constructed by retrieving the text and image candidates with the highest cosine similarity to each query.
We evaluate our method across seven VLMs using two multi-modal QA benchmarks~(MMQA and WebQA), where each question requires retrieving either text or image data.
Our experimental results show that our method significantly improves retrieval performance, achieving average Recall@20 gains of 64\% on MMQA and 28\% on WebQA when the query and the target data belong to different modalities.
Compared to E5-V, which addresses the modality gap through image captioning, we confirm that our method more effectively bridges the modality gap.
\end{abstract}

\section{Introduction}
Information retrieval (IR) plays a key role in a wide range of NLP applications, including web search engines~\cite{Kobayashi2000} and question answering systems~\cite{Kolomiyets2011}.
While traditional approaches primarily focus on retrieving textual information~\cite{Robertson2009,Karpukhin2020}, there is a growing interest in retrieving both text and images to provide richer and more informative results~\cite{Zhou2024_MARVEL}.

\begin{figure}[t]
  \centering
  \includegraphics[width=0.9\linewidth]{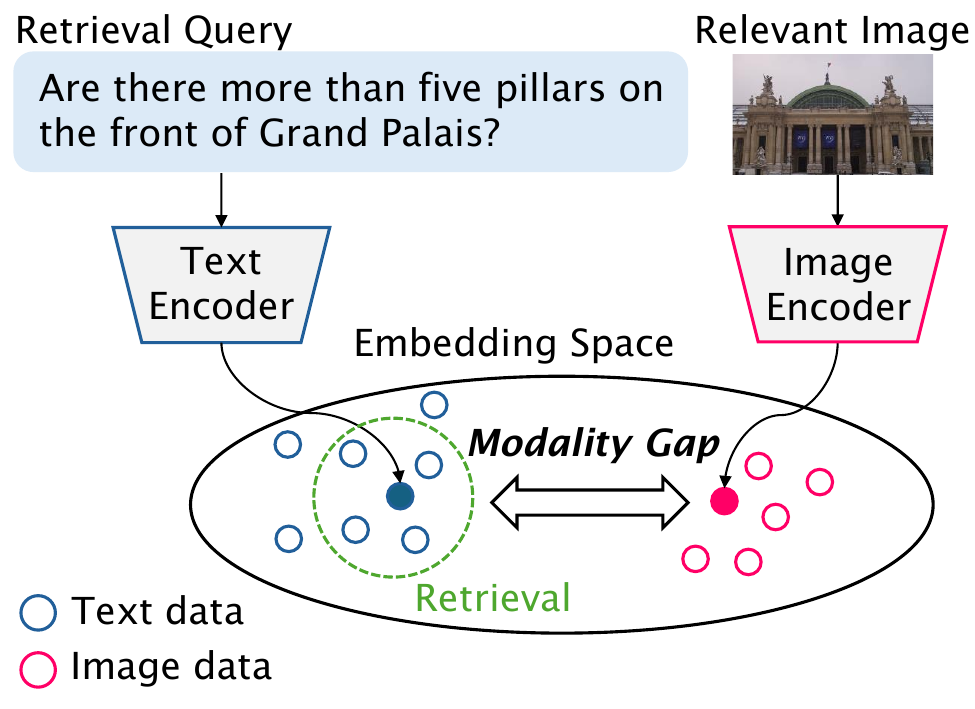}
  \caption{Conceptual overview of the modality gap. Texts and their corresponding images are projected to distant regions of the embedding space.}
  \label{fig: motivation}
\end{figure}

Vision-language models~(VLMs), such as CLIP~\cite{Radford2021}, enable both text and image data to be embedded into a shared representation space.
Although VLMs enable effective text-to-image retrieval~\cite{Radford2021}, it is still challenging to extract relevant information from a database that contains both text and images.
Specifically, text items often dominate the top-ranked results even when relevant images exist~\cite{Chang2021,Liu2023}.
This issue is attributed to the \textit{modality gap}---a phenomenon in which embeddings from different modalities are mapped to separate regions of the representation space~\cite{Liang2022}.
Consequently, data that share the same modality as the query tend to receive higher similarity scores, regardless of actual relevance~(illustrated in Figure~\ref{fig: motivation}).

To address this problem, several approaches have been proposed.
Some methods address the modality gap by fine-tuning pre-trained VLMs using paired datasets consisting of queries and their manually labeled corresponding text or image data~\cite{Fahim2024,Eslami2025}.
Other methods for converting visual data into text have also been introduced, such as E5-V~\cite{Jiang2024E5V}.
However, these approaches have shortcomings: collecting human-annotated data is resource-intensive, whereas image captioning would fail to preserve necessary visual information in text.

In this study, we propose a retrieval method that mitigates the impact of modality gap without manually labeled data or image captioning.
The key idea is to make similarity scores comparable across modalities by standardizing them using the modality-specific mean and variance.
To estimate these statistics, we construct pseudo-positive pairs of unlabeled queries and their most similar texts or images.
We then derive modality-specific mean and variance from these pairs, which are used to standardize similarity scores during retrieval.

To evaluate our approach, we conduct experiments on multi-modal question answering benchmarks, i.e., MMQA~\cite{talmor2021} and \mbox{Web}QA~\cite{Chang2021} with seven pre-trained VLMs.
Our method significantly improves retrieval performance when the query and the target data belong to different modalities, achieving average gains of 64\% and 28\% in Recall@20 on MMQA and WebQA, respectively.

Our main contributions are as follows:
\begin{itemize}[]
  \item We propose a similarity standardization approach to mitigate the effect of the modality gap on multi-modal retrieval.
  \item Our method improves the retrieval performances on two datasets regardless of modalities, compared to E5-V.
  \item Our method bridges the modality gap without manually labeled datasets, such as pairs of queries and their corresponding examples.
\end{itemize}

\begin{figure*}[t]
  \centering
  \includegraphics[width=0.95\linewidth]{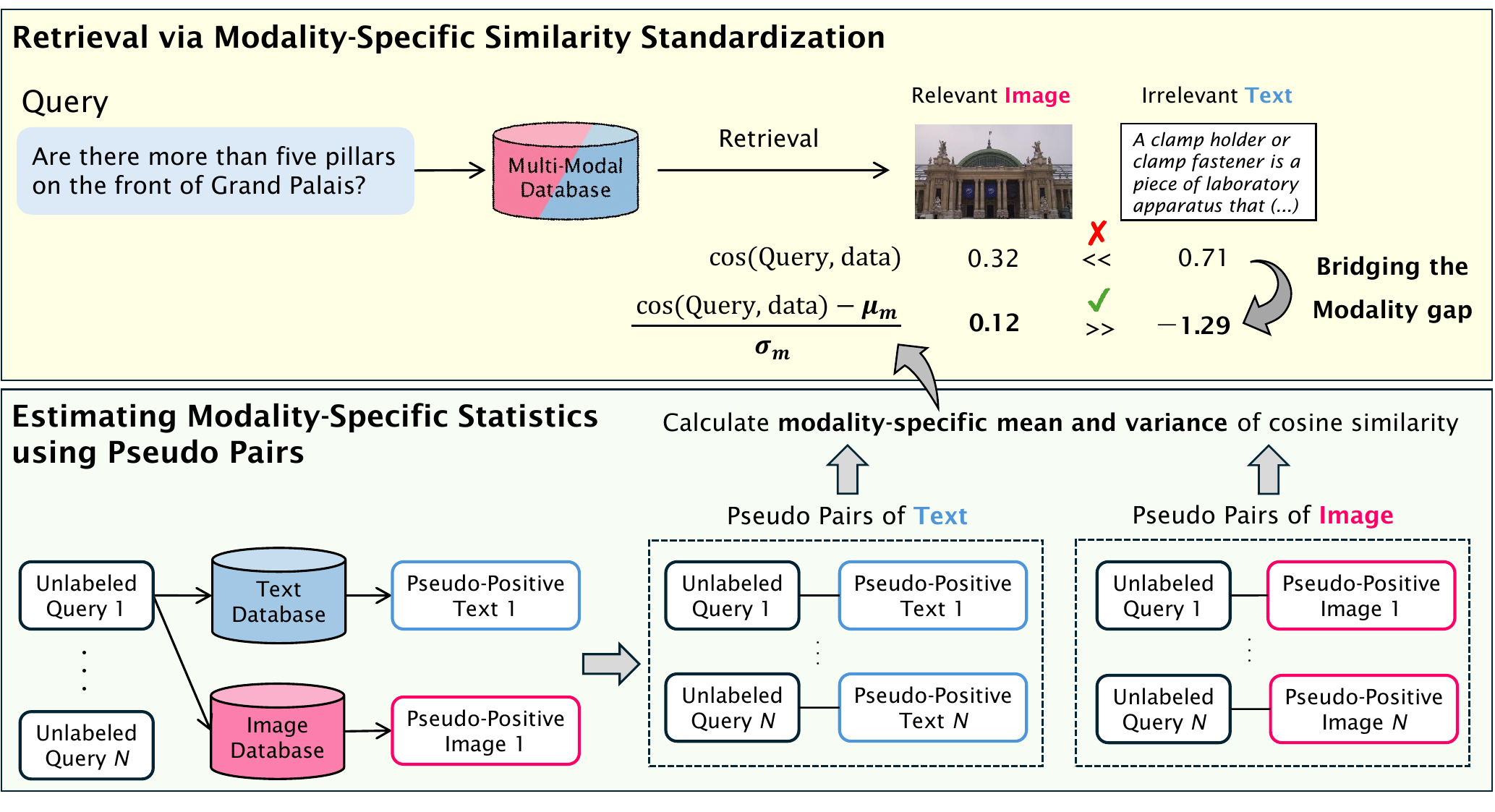}
  \caption{Overview of our proposed method. The modality gap causes irrelevant text to score higher than relevant images. Our approach addresses this issue by standardizing cosine similarity scores based on modality-specific mean and variance calculated from pseudo data.}
  \label{fig: overview}
\end{figure*}
\section{Related Work}
\subsection{Multi-Modal Retrieval}
Vision-language models~(VLMs) have shown remarkable progress in recent years~\cite{Radford2021,Jia2021}.
These models are typically trained using contrastive learning to align images and text in a representation space.
Their embeddings can be used for retrieval by computing similarity scores with each item in the database~\cite{Karpukhin2020}.

Retrieval tasks involving multiple modalities can be broadly categorized into two settings~\cite{Liu2023}.
\textit{Cross modality retrieval} refers to settings in which the query and target belong to different modalities, such as text-image or image-text retrieval.
In contrast, \textit{multi-modal retrieval} assumes that the retrieval database contains data from multiple modalities---for example, both text and images---and the goal is to find the most relevant item regardless of its modality.

While contrastively trained VLMs perform well in cross modality retrieval tasks~\cite{Radford2021}, their performance in multi-modal retrieval remains limited.
In particular, when both text and images are present in the retrieval set, these models often retrieve items only from the same modality as the query, and fail to retrieve relevant data from the other modality~\cite{Chang2021,cohere_multimodal_embed3_2024}.

This issue is attributed to the modality gap, a clear separation between image and text embeddings of contrastively trained VLMs.
This phenomenon was first studied by \citet{Liang2022}, who showed that it exists even in randomly initialized models and persists throughout contrastive training.
Several causes have been suggested in prior work, including an information imbalance between text and image inputs~\cite{Schrodi2025}.

\subsection{Bridging the Modality Gap}
Some approaches attempt to eliminate the modality gap in VLMs by modifying the contrastive training process.
\cite{Fahim2024} augment CLIP's objective with uniformity and alignment regularizers to enforce balanced embedding distributions and eliminate the modality gap.
\citet{Schrodi2025} demonstrated that contrastive learning can mitigate the modality gap when the training data is balanced in information content across modalities.
\citet{Eslami2025} introduce AlignCLIP, which adds shared parameters between visual and text encoders and an intra-modality separation term to the contrastive loss.
While effective, these methods require access to manually paired datasets, which can be expensive or unavailable in real-world scenarios.

Another line of work obtains image embeddings by leveraging image captions~\cite{Liu2023,Zhou2024_VISTA, Zhou2024_MARVEL}.
These models achieve strong performance in multi-modal retrieval, but rely heavily on captions.
In settings without image descriptions, retrieval quality deteriorates, indicating limited use of visual features.

More recently, methods utilizing the vision-language capabilities of multi-modal large language models~(MLLMs) have been explored~\cite{Jiang2024E5V,zhang2024gme,MMEmbed}.
For instance, E5-V~\cite{Jiang2024E5V} prompts its backbone MLLM with an image to generate a one-word summary of it.
By using the resulting features to obtain image embeddings, E5-V aligns visual inputs with the language space, effectively eliminating the modality gap.

Unlike the existing works that require manually labeled data or image captioning, our method directly adjusts similarity scores across modalities using pseudo-positive examples, eliminating the need for manual supervision.
\section{Task Formulation}
\label{sec: task}
We work on the task of retrieving relevant data from a multi-modal database that contains both text and images, given a natural language query.

Formally, let $q$ be a textual query and let $\mathcal{D}=\mathcal{D}_{\text{text}} \cup \mathcal{D}_{\text{image}}$ denote the retrieval database, where  $\mathcal{D}_{\text{text}}$ and $\mathcal{D}_{\text{image}}$ are sets of textual and visual items respectively.
A pre-trained VLM $f$ encodes both the query and each item in the database into the same space.
For each candidate $d \in \mathcal{D}$, its relevance to the query can be measured by comparing their embeddings, for example, using cosine similarity: $\cos(f(q), f(d))$.

However, due to the modality gap, similarities differ in scale between text and image modalities.
Specifically, a text query tends to assign higher scores to textual candidates than to images, causing relevant images to appear lower in the ranking.
\section{Proposed Methods}
\label{sec: methods}
In this section, we propose a method that mitigates the negative impact of the modality gap without manually labeled data.
We first introduce similarity standardization approach as described in Section~\ref{subsec: standardization}.
Then, we construct pseudo pairs instead of labeled data, as detailed in Section~\ref{subsec: pseudo_positive}.

\subsection{Modality-Specific Similarity Standardization}
\label{subsec: standardization}
To bridge the modality gap, we propose a similarity standardization approach with modality-specific statistics.
We standardize the similarity scores between queries and target information (i.e., positive examples) using their means and variances computed separately for text targets and image targets.

Let $\mathcal{P}_m$ be a set of query-positive pairs where positive example belongs to modality $m \in \{\text{text}, \text{image}\}$.
We calculate the mean and variance of similarities for each modality as:
\begin{align}
  \label{eq:statistics}
  \mu_m &= \frac{1}{|\mathcal{P}_m|}\sum_{(q,d^+_m) \in \mathcal{P}_m}\cos(f(q), f(d^+_m)), \notag \\
  \sigma_{m}^2 &= \frac{1}{|\mathcal{P}_m|}\sum_{(q,d^+_m) \in \mathcal{P}_m}(\cos(f(q), f(d^+_m)) - \mu_m)^2,
\end{align}
where each $(q, d^+_m)\in \mathcal{P}_m$ is a query-positive pair.

Using the modality-specific statistics estimated above, we standardize the cosine similarity between a query $q$ and a candidate $d \in \mathcal{D}$ of modality $m$ as:
\begin{equation}
  \text{sim}(q, d) = \frac{\cos(f(q), f(d)) - \mu_m}{\sigma_m}.
\end{equation}
This modality-aware standardization will mitigate the negative impact of the modality gap on similarities between text and image.
Note that the statistics $\mu_m$ and $\sigma_{m}^2$ are computed from the pre-collected dataset $\mathcal{P}_m$ and remain fixed regardless of the retrieval queries.

\subsection{Pseudo Pair Construction}
\label{subsec: pseudo_positive}
We propose a method for constructing pseudo data that eliminates the need for manually labeled data.

Let $\mathcal{D}_m$ be the subset of the retrieval database corresponding to modality $m \in \{\text{text}, \text{image}\}$, and let $\mathcal{Q}$ denote a set of unlabeled queries.  
Given a query $q \in \mathcal{Q}$, we extract the most similar item from $\mathcal{D}_m$ for each modality $m$, and treat it as a pseudo-positive example of modality $m$:
\begin{equation}
  \label{eq:sample_pseudo_positive}
  \hat{d}_m^+ = \argmax_{d \in \mathcal{D}_m} \cos(f(q), f(d)).
\end{equation}
By repeating this process for all queries in $\mathcal{Q}$, we construct a modality-specific pseudo pair set $\hat{\mathcal{P}}_m$ for each modality $m$:
\begin{equation}
  \hat{\mathcal{P}_m} = \{(q, \hat{d^+_m}) \mid q \in \mathcal{Q}\}.
\end{equation}
$\hat{\mathcal{P}}_m$ can be used as a substitute for the manually labeled set $\mathcal{P}_m$ in Equations~(\ref{eq:statistics}).
This allows our method to perform modality-specific standardization without relying on any labeled data.
\section{Experimental Setup}
\subsection{Datasets for Evaluation}
We evaluate our method on two multi-modal question answering datasets: MultimodalQA~\cite{talmor2021} and WebQA~\cite{Chang2021}.
These datasets are widely used benchmarks for the multi-modal retrieval task~\cite{Chen2022MuRAG,Liu2023,Zhou2024_VISTA,Zhou2024_MARVEL}.
In our experiments, we use questions that require retrieving relevant textual passages~(TextQ) or images~(ImageQ) in order to answer them.
Table~\ref{table: dataset_example} shows examples from each dataset, and Table~\ref{table: dataset_statistics} shows the dataset sizes.

\noindent
\textbf{MultiModalQA~(MMQA)}~\citep{talmor2021} is a benchmark for multi-hop question answering across multiple modalities, including text, images, and tables. 
It is constructed from Wikipedia tables linked with relevant textual paragraphs and images via shared entities. 

\noindent
\textbf{WebQA}~\cite{Chang2021} is a large-scale open-domain question answering dataset that includes questions paired with corresponding textual passages or images.
The data is collected from the open web and Wikipedia.
Following \citet{Liu2023} and \citet{Zhou2024_MARVEL}, we construct a retrieval corpus by collecting all images and text passages relevant to all queries in the WebQA dataset.

\begin{table*}[t]
  \centering
  \small
  \begin{tabular}{llp{5cm}p{7cm}}
    \toprule
    Dataset               & Type                   & Question & Positive Example \\
    \midrule
    \multirow{2}{*}{MMQA} & TextQ  & When did ``Harry Potter and the Sorcerer's Stone'' movie come out? & Harry Potter and the Philosopher's Stone (released in the United States as Harry Potter and the Sorcerer's Stone) is a 2001 fantasy film directed by Chris Columbus and distributed by Warner Bros.\\
    \cmidrule(lr){2-4}
    & ImageQ & How many colors are on the Mississippi flag? & Refer to Figure~\ref{fig: MMQA_img}. \\
    \cmidrule(lr){1-4}
    \multirow{2}{*}{WebQA} & TextQ & What part of the human body does the nerves in the frontalis muscle serve and the occipitofrontalis muscle serve? & The frontalis muscle is supplied by the facial nerve and receives blood from the supraorbital and supratrochlear arteries. In humans, the occipitofrontalis only serves for facial expressions.\\
    \cmidrule(lr){2-4}
    & ImageQ & Are there more than five pillars on the front of Grand Palais? & Refer to Figure~\ref{fig: WebQA_img}. \\
    \bottomrule
  \end{tabular}
  \caption{Examples from MMQA and WebQA datasets. Each dataset includes two types of questions: TextQ and ImageQ, which refer to questions that require retrieving text and images to answer, respectively.}
  \label{table: dataset_example}
\end{table*}

\begin{figure}[t]
  \centering
  \begin{subfigure}{0.45\columnwidth}
    \centering
    \includegraphics[height=2cm]{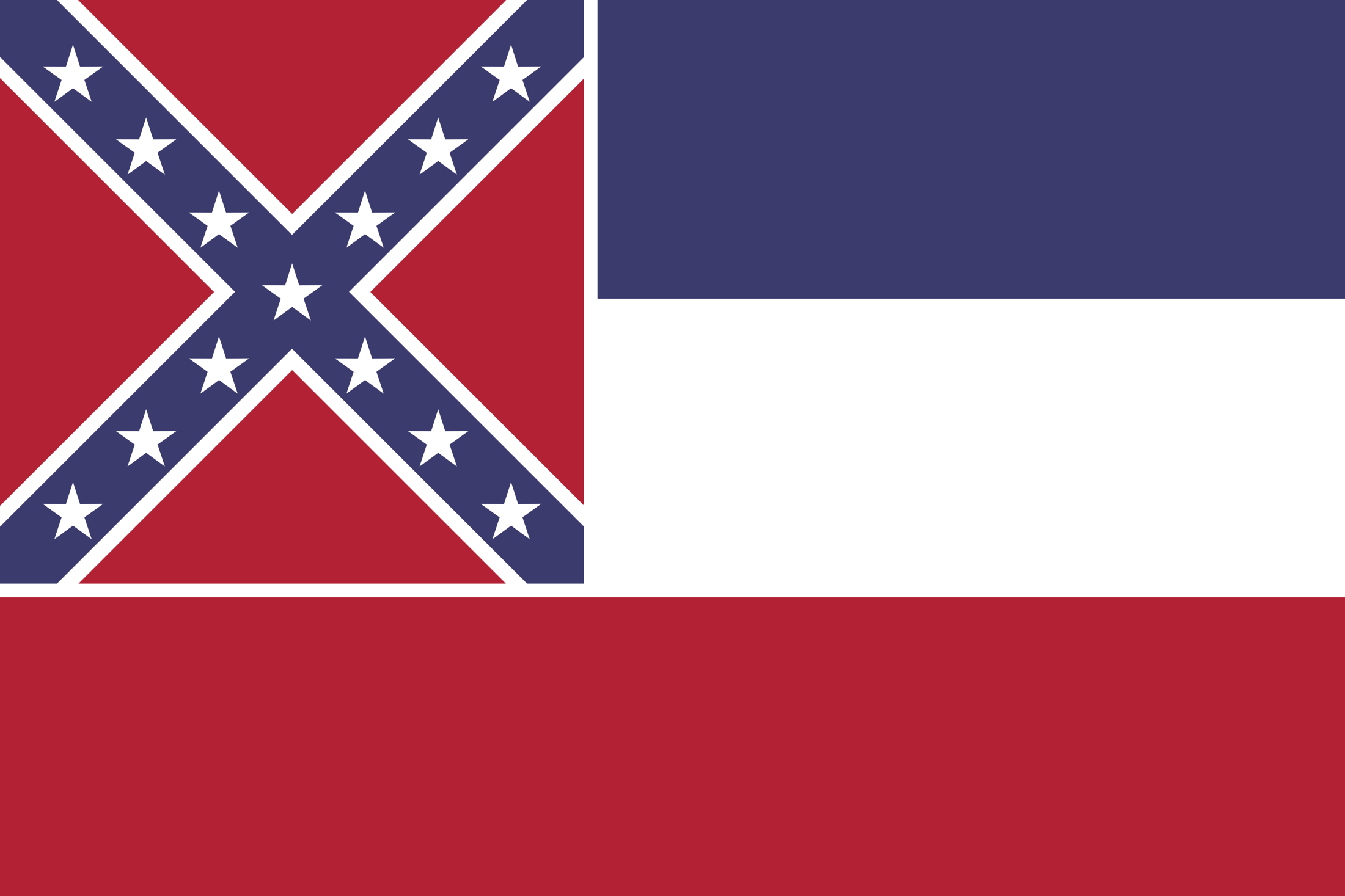}
    \caption{MMQA: ``\textit{How many colors are on the Mississippi flag?}''}
    \label{fig: MMQA_img}
  \end{subfigure}
  \hspace{5mm}
  \begin{subfigure}{0.45\columnwidth}
    \centering
    \includegraphics[height=2cm]{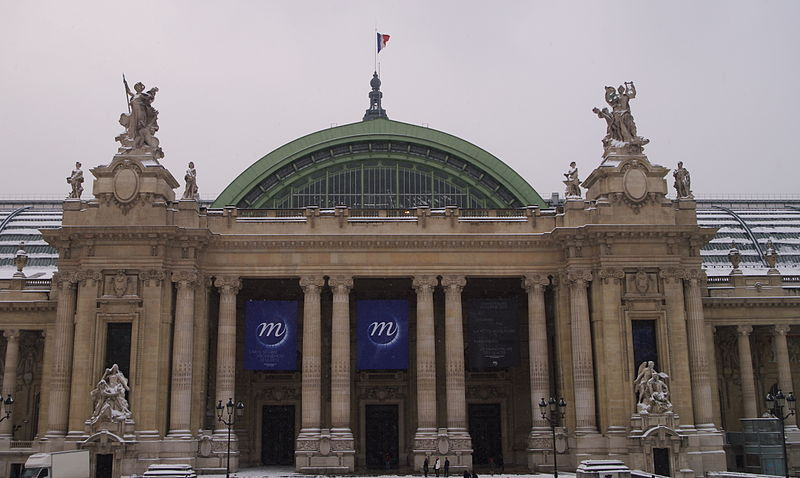}
    \caption{WebQA: ``\textit{Are there more than five pillars on the front of Grand Palais?}''}
    \label{fig: WebQA_img}
  \end{subfigure}
  \caption{Examples of positive images for ImageQ in MMQA and WebQA shown in Table~\ref{table: dataset_example}.}
\end{figure}
\begin{table}[t]
  \centering
  \small
  \begin{tabular}{lrrrr}
    \toprule
    \multirow{2}{*}{\# of dataset} & \multicolumn{2}{c}{Source} & \multicolumn{2}{c}{Query} \\
    \cmidrule(lr){2-3} \cmidrule(lr){4-5}
                             & text & image & TextQ & ImageQ \\
    \midrule
    MMQA  & 218K & 57K  & 6.7K/721 & 1.9K/230 \\
    WebQA & 787K & 389K & 15K/2.4K & 16K/2.5K \\
    \bottomrule
  \end{tabular}
  \caption{Numbers of retrieval candidates and queries in MMQA and WebQA. The numbers of queries are listed as training/test. Validation data is not used in our experiments.}
  \label{table: dataset_statistics}
\end{table}

\subsection{Datasets for Pseudo Pair Construction}
Pseudo pairs are constructed independently for the MMQA and WebQA datasets.
We use queries from the training split of each dataset and sample their pseudo-positive examples from the retrieval source of each dataset as illustrated in Equation~\ref{eq:sample_pseudo_positive}.

\subsection{Metrics}
We evaluate our methods using Recall@$k$, MRR@$k$, and NDCG@$k$.
All metrics are primarily measured at $k=20$.
For Recall, we additionally compute values at $k{=}1$, $5$, and $100$ to examine the effect of varying $k$.

\subsection{Models}
We apply our method to seven pre-trained VLMs to demonstrate its robust effectiveness.
To assess models expected to exhibit a modality gap due to contrastive training, we include CLIP~\cite{Radford2021}~(ViT-B/32 and ViT-L/14), Long-CLIP~\cite{Zhang2024LongCLIP}~(base and large), and BLIP~\cite{Li2022BLIP}.
We also include Cohere Embed 3 English~\cite{cohere_multimodal_embed3_2024}, a high-performance VLM accessible via API.  
In addition, we evaluate E5-V~\cite{Jiang2024E5V}, which integrates image captioning via a MLLM.
While E5-V is designed to mitigate the modality gap, we apply similarity standardization to examine whether our method can further improve its performance.
The computational resources are provided in Appendix~\ref{subsec: resources}.

\subsection{Evaluation Conditions}
All VLMs are evaluated under the following three configurations. \\
(i) \textbf{Cos}: Cosine similarities are simply used for retrieval. \\
(ii) \textbf{Std}: Cosine similarities are standardized by our method with manually labeled data, which is taken from the training split of each dataset. \\
(iii) \textbf{Ours}: Cosine similarities are standardized by our method with our pseudo pairs.
\section{Results and Discussions}
\label{sec: results}
\subsection{Overall Results}
Table~\ref{table: overall_result} summarizes the overall retrieval performance across seven VLMs on MMQA and WebQA datasets.
When Cos was applied, four of the CLIP-based models and BLIP retrieved almost no relevant results, resulting in near-zero scores on all evaluation metrics on ImageQ.
This suggests that the modality gap causes irrelevant text passages to be ranked higher than relevant images, hindering accurate retrieval.

In contrast, applying our method to these models significantly improved the performances, achieving average gains of 64\% and 28\% in Recall@$20$ for MMQA ImageQ and WebQA ImageQ, respectively, thereby confirming its effectiveness in bridging the modality gap.

Notably, all models with our method outperformed E5-V on ImageQ.
These results highlight the advantage of processing images without any loss of information, different from the existing works with image captioning or verbalization.
Although a slight performance degradation was observed on TextQ, the overall trade-off is favorable with notable gains on ImageQ.

Cohere Embed 3 and E5-V achieved high performance on TextQ, with approximately 80\% in Recall@20.
On ImageQ, they retained a certain level of performance without our method, achieving Recall@20 ranging from 40-50\% on MMQA and 10-20\% on WebQA.
For E5-V, this can be attributed to its strong capability for understanding textual information through its MLLM backbone, as well as its architecture that converts images into text.
While the architecture and training details of Cohere Embed 3 are not publicly available, its performance suggests that it may adopt a similar architecture or training process to models like E5-V.
When our standardization is applied to these models, further improvements are observed on ImageQ; however, it also results in a large drop in TextQ accuracy compared to CLIP-based models and BLIP.
This indicates that the benefit of our method is limited when the modality gap is already small.

\begin{table*}[t]
  \centering
  \small
  \resizebox{\textwidth}{!}{
  \begin{tabular}{llrrrrrrrrrrrr}
    \toprule
    \multirow{3}{*}{Model} & \multirow{3}{*}{Method} & \multicolumn{6}{c}{MMQA} & \multicolumn{6}{c}{WebQA} \\
    & & \multicolumn{3}{c}{TextQ} & \multicolumn{3}{c}{ImageQ} & \multicolumn{3}{c}{TextQ} & \multicolumn{3}{c}{ImageQ} \\
    \cmidrule(lr){3-5} \cmidrule(lr){6-8} \cmidrule(lr){9-11} \cmidrule(lr){12-14}
    & & Recall & MRR & NDCG & Recall & MRR & NDCG & Recall & MRR & NDCG & Recall & MRR & NDCG  \\
    \midrule
    \multirow{3}{*}{CLIP~(ViT-B/32)} & Cos & $31.90$ & $26.62$ & $23.90$ & $0.00$ & $0.00$ & $0.00$ & $28.89$ & $21.14$ & $18.89$ & $0.00$ & $0.00$ & $0.00$ \\
                                     & Std & $31.55$ & $25.78$ & $23.25$ & $52.61$ & $36.65$ & $40.32$ & $23.96$ & $16.38$ & $15.01$ & $32.82$ & $15.20$ & $18.02$ \\
                                     & Ours & \cellcolor{cyan!20}$27.46$ & \cellcolor{cyan!20}$18.44$ & \cellcolor{cyan!20}$17.88$ & \cellcolor{cyan!20}$66.09$ & \cellcolor{cyan!20}$45.03$ & \cellcolor{cyan!20}$49.86$ & \cellcolor{cyan!20}$27.14$ & \cellcolor{cyan!20}$19.07$ & \cellcolor{cyan!20}$17.29$ & \cellcolor{cyan!20}$28.14$ & \cellcolor{cyan!20}$13.79$ & \cellcolor{cyan!20}$15.98$ \\
    \cmidrule{1-14}
    \multirow{3}{*}{CLIP~(ViT-L/14)} & Cos  & $35.51$ & $28.63$ & $25.86$ & $1.30$ & $0.41$ & $0.62$ & $32.60$ & $24.05$ & $21.45$ & $0.04$ & $0.00$ & $0.01$ \\
                                     & Std  & $35.37$ & $27.60$ & $25.16$ & $62.17$ & $43.32$ & $47.70$ & $28.84$ & $19.37$ & $17.91$ & $43.55$ & $22.54$ & $25.76$ \\
                                     & Ours & \cellcolor{cyan!20}$31.28$ & \cellcolor{cyan!20}$21.48$ & \cellcolor{cyan!20}$20.54$ & \cellcolor{cyan!20}$76.52$ & \cellcolor{cyan!20}$58.88$ & \cellcolor{cyan!20}$63.05$ & \cellcolor{cyan!20}$31.27$ & \cellcolor{cyan!20}$21.44$ & \cellcolor{cyan!20}$19.69$ & \cellcolor{cyan!20}$37.10$ & \cellcolor{cyan!20}$20.37$ & \cellcolor{cyan!20}$22.75$ \\
    \cmidrule{1-14}
    \multirow{3}{*}{Long-CLIP-B} & Cos  & $58.67$ & $45.11$ & $43.02$ & $0.00$ & $0.00$ & $0.00$ & $43.93$ & $30.92$ & $28.56$ & $0.00$ & $0.00$ & $0.00$ \\
                                 & Std  & $54.65$ & $40.73$ & $38.76$ & $66.09$ & $47.67$ & $51.94$ & $34.94$ & $23.79$ & $22.05$ & $33.01$ & $14.92$ & $17.98$ \\
                                 & Ours & \cellcolor{cyan!20}$53.33$ & \cellcolor{cyan!20}$35.48$ & \cellcolor{cyan!20}$35.04$ & \cellcolor{cyan!20}$66.96$ & \cellcolor{cyan!20}$50.72$ & \cellcolor{cyan!20}$54.51$ & \cellcolor{cyan!20}$40.44$ & \cellcolor{cyan!20}$27.72$ & \cellcolor{cyan!20}$25.79$ & \cellcolor{cyan!20}$28.59$ & \cellcolor{cyan!20}$13.43$ & \cellcolor{cyan!20}$15.97$ \\
    \cmidrule{1-14}
    \multirow{3}{*}{Long-CLIP-L} & Cos  & $63.04$ & $45.56$ & $44.32$ & $0.43$ & $0.11$ & $0.19$ & $45.18$ & $30.36$ & $28.54$ & $0.00$ & $0.00$ & $0.00$ \\
                                 & Std  & $58.39$ & $41.94$ & $40.70$ & $71.74$ & $49.38$ & $54.52$ & $35.66$ & $23.57$ & $22.14$ & $39.84$ & $20.22$ & $23.36$ \\
                                 & Ours & \cellcolor{cyan!20}$57.07$ & \cellcolor{cyan!20}$38.60$ & \cellcolor{cyan!20}$38.19$ & \cellcolor{cyan!20}$73.91$ & \cellcolor{cyan!20}$54.04$ & \cellcolor{cyan!20}$58.63$ & \cellcolor{cyan!20}$41.46$ & \cellcolor{cyan!20}$27.14$ & \cellcolor{cyan!20}$25.69$ & \cellcolor{cyan!20}$35.34$ & \cellcolor{cyan!20}$18.38$ & \cellcolor{cyan!20}$21.09$ \\
    \cmidrule{1-14}
    \multirow{3}{*}{BLIP} & Cos  & $41.75$ & $30.20$ & $28.64$ & $0.00$ & $0.00$ & $0.00$ & $37.15$ & $27.07$ & $24.23$ & $0.00$ & $0.00$ & $0.00$ \\
                          & Std  & $40.92$ & $28.58$ & $27.42$ & $39.57$ & $23.97$ & $27.54$ & $24.00$ & $14.75$ & $14.04$ & $17.62$ & $8.24$ & $9.73$ \\
                          & Ours & \cellcolor{cyan!20}$36.75$ & \cellcolor{cyan!20}$23.33$ & \cellcolor{cyan!20}$23.31$ & \cellcolor{cyan!20}$43.48$ & \cellcolor{cyan!20}$27.45$ & \cellcolor{cyan!20}$31.15$ & \cellcolor{cyan!20}$31.40$ & \cellcolor{cyan!20}$20.71$ & \cellcolor{cyan!20}$19.23$ & \cellcolor{cyan!20}$14.04$ & \cellcolor{cyan!20}$6.35$ & \cellcolor{cyan!20}$7.62$ \\
    \cmidrule{1-14}
    \multirow{3}{*}{Cohere Embed 3} & Cos  & $87.17$ & $78.81$ & $74.72$ & $50.43$ & $20.79$ & $27.61$ & $76.52$ & $59.19$ & $55.86$ & $20.43$ & $8.00$ & $10.16$ \\
                                    & Std  & $72.19$ & $66.33$ & $60.63$ & $52.17$ & $27.24$ & $32.92$ & $54.78$ & $41.69$ & $38.19$ & $27.42$ & $12.36$ & $14.83$ \\
                                    & Ours & \cellcolor{cyan!20}$73.99$ & \cellcolor{cyan!20}$63.25$ & \cellcolor{cyan!20}$59.20$ & \cellcolor{cyan!20}$52.17$ & \cellcolor{cyan!20}$28.17$ & \cellcolor{cyan!20}$33.61$ & \cellcolor{cyan!20}$69.23$ & \cellcolor{cyan!20}$52.67$ & \cellcolor{cyan!20}$49.36$ & \cellcolor{cyan!20}$25.39$ & \cellcolor{cyan!20}$11.48$ & \cellcolor{cyan!20}$13.73$ \\
    \cmidrule{1-14}
    \multirow{3}{*}{E5-V} & Cos  & $84.88$ & $66.67$ & $67.20$ & $38.70$ & $17.34$ & $22.06$ & $74.37$ & $54.88$ & $52.27$ & $11.89$ & $5.19$ & $6.37$ \\
                          & Std  & $80.79$ & $63.33$ & $63.56$ & $41.74$ & $21.33$ & $25.91$ & $48.61$ & $35.04$ & $33.11$ & $21.05$ & $9.75$ & $11.50$ \\
                          & Ours & \cellcolor{cyan!20}$70.39$ & \cellcolor{cyan!20}$53.15$ & \cellcolor{cyan!20}$53.12$ & \cellcolor{cyan!20}$41.74$ & \cellcolor{cyan!20}$21.55$ & \cellcolor{cyan!20}$26.09$ & \cellcolor{cyan!20}$65.73$ & \cellcolor{cyan!20}$48.76$ & \cellcolor{cyan!20}$46.11$ & \cellcolor{cyan!20}$18.78$ & \cellcolor{cyan!20}$8.87$ & \cellcolor{cyan!20}$8.87$ \\
    \bottomrule
  \end{tabular}
  }
  \caption{Overall retrieval results on MMQA and WebQA. Recall@20, MRR@20, and NDCG@20 are reported. Cos uses cosine similarity as the retrieval score. Std and Ours apply similarity standardization using modality-specific mean and variance estimated from labeled and pseudo pairs, respectively.}
  \label{table: overall_result}
\end{table*}

\subsection{Severe Impact of the Modality Gap}
To examine how the modality gap affects retrieval performance, we evaluated Recall at various cutoff values of retrieval on ImageQ.
Table~\ref{table: various_recall} reports Recall@\{1, 5, 20, 100\} for each model and dataset.

For Cos, increasing the number of retrieved candidates had almost no effect---Recall@$k$ remained around zero even with $k=100$.
This result clearly indicates that the modality gap severely degrades retrieval performance on ImageQ.

In contrast, our method yields substantial improvements in Recall@$k$ across all tested values of $k$, demonstrating its effectiveness in bridging the modality gap.

\begin{table*}
  \centering
  \small
  \begin{tabular}{llrrrrrrrr}
    \toprule
    \multirow{2}{*}{Model} & \multirow{2}{*}{Method} & \multicolumn{4}{c}{MMQA} & \multicolumn{4}{c}{WebQA} \\
    \cmidrule(lr){3-6} \cmidrule(lr){7-10}
    & & 1 & 5 & 20 & 100 & 1 & 5 & 20 & 100 \\
    \midrule
    \multirow{2}{*}{CLIP~(ViT-B/32)} & Cos & 0.00 & 0.00 & 0.00 & 0.00 & 0.00 & 0.00 & 0.00 & 0.00 \\
                                     &Ours & \cellcolor{cyan!20}37.39 & \cellcolor{cyan!20}53.48 & \cellcolor{cyan!20}66.09 & \cellcolor{cyan!20}72.17 & \cellcolor{cyan!20}8.16 & \cellcolor{cyan!20}16.11 & \cellcolor{cyan!20}28.14 & \cellcolor{cyan!20}44.78 \\
    \cmidrule{1-10}
    \multirow{2}{*}{CLIP~(ViT-L/14)} & Cos & 0.00 & 0.87 & 1.30 & 3.04 & 0.00 & 0.00 & 0.04 & 0.06 \\
                                     & Ours& \cellcolor{cyan!20}50.87 & \cellcolor{cyan!20}69.57 & \cellcolor{cyan!20}76.52 & \cellcolor{cyan!20}81.74 & \cellcolor{cyan!20}12.90 & \cellcolor{cyan!20}24.39 & \cellcolor{cyan!20}37.10 & \cellcolor{cyan!20}54.76 \\
    \cmidrule{1-10}
    \multirow{2}{*}{Long-CLIP-B}     & Cos & 0.00 & 0.00 & 0.00 & 0.00 & 0.00 & 0.00 & 0.00 & 0.00 \\
                                     & Ours& \cellcolor{cyan!20}43.91 & \cellcolor{cyan!20}60.43 & \cellcolor{cyan!20}66.96 & \cellcolor{cyan!20}76.96 & \cellcolor{cyan!20}7.89 & \cellcolor{cyan!20}16.21 & \cellcolor{cyan!20}28.59 & \cellcolor{cyan!20}45.46 \\
    \cmidrule{1-10}
    \multirow{2}{*}{Long-CLIP-L}     & Cos & 0.00 & 0.43 & 0.43 & 0.87 & 0.00 & 0.00 & 0.00 & 0.00 \\
                                     & Ours& \cellcolor{cyan!20}46.09 & \cellcolor{cyan!20}63.48 & \cellcolor{cyan!20}73.91 & \cellcolor{cyan!20}80.87 & \cellcolor{cyan!20}11.95 & \cellcolor{cyan!20}21.39 & \cellcolor{cyan!20}35.34 & \cellcolor{cyan!20}52.77 \\
    \cmidrule{1-10}
    \multirow{2}{*}{BLIP} & Cos & 0.00  & 0.00  & 0.00  & 0.00  & 0.00 & 0.00 & 0.00  & 0.00 \\
                          & Ours& \cellcolor{cyan!20}21.30 & \cellcolor{cyan!20}35.22 & \cellcolor{cyan!20}43.48 & \cellcolor{cyan!20}56.09 & \cellcolor{cyan!20}3.78 & \cellcolor{cyan!20}7.39 & \cellcolor{cyan!20}14.04 & \cellcolor{cyan!20}26.66 \\
    \cmidrule{1-10}
    \multirow{2}{*}{Cohere Embed 3} & Cos & 10.00 & 34.78 & 50.43 & 64.78 & 4.38 & 9.14  & 20.43 & 40.60 \\
                                    & Ours& \cellcolor{cyan!20}20.43 & \cellcolor{cyan!20}38.26 & \cellcolor{cyan!20}52.17 & \cellcolor{cyan!20}65.65 & \cellcolor{cyan!20}6.41 & \cellcolor{cyan!20}13.52 & \cellcolor{cyan!20}25.39 & \cellcolor{cyan!20}43.83 \\
    \cmidrule{1-10}
    \multirow{2}{*}{E5-V} & Cos & 12.17 & 23.91 & 38.70 & 59.57 & 2.95 & 6.35  & 11.89 & 26.52 \\
                          & Ours& \cellcolor{cyan!20}16.09 & \cellcolor{cyan!20}28.26 & \cellcolor{cyan!20}41.74 & \cellcolor{cyan!20}63.04 & \cellcolor{cyan!20}5.10 & \cellcolor{cyan!20}10.49 & \cellcolor{cyan!20}18.78 & \cellcolor{cyan!20}36.80 \\
    \bottomrule
  \end{tabular}
  \caption{Results of Recall@$k$~($k = \{1, 5, 20, 100\}$) for each model on ImageQ queries in MMQA and WebQA datasets.}
  \label{table: various_recall}
\end{table*}

\subsection{Pseudo Pairs vs. Manually Labeled Pairs}
To assess how pseudo pairs affect retrieval, we compared retrieval performances of the Std method and our method.
Table~\ref{table: overall_result} shows that the results of our method were equal or higher than those of the Std method.
This result demonstrates that pseudo pairs can serve as an effective substitute for manually labeled pairs.
\section{Analysis of the Modality Gap}
\subsection{The Effect of Standardization}
\begin{figure}[t]
  \centering
  \begin{minipage}{0.45\columnwidth}
    \centering
    \includegraphics[width=\columnwidth]{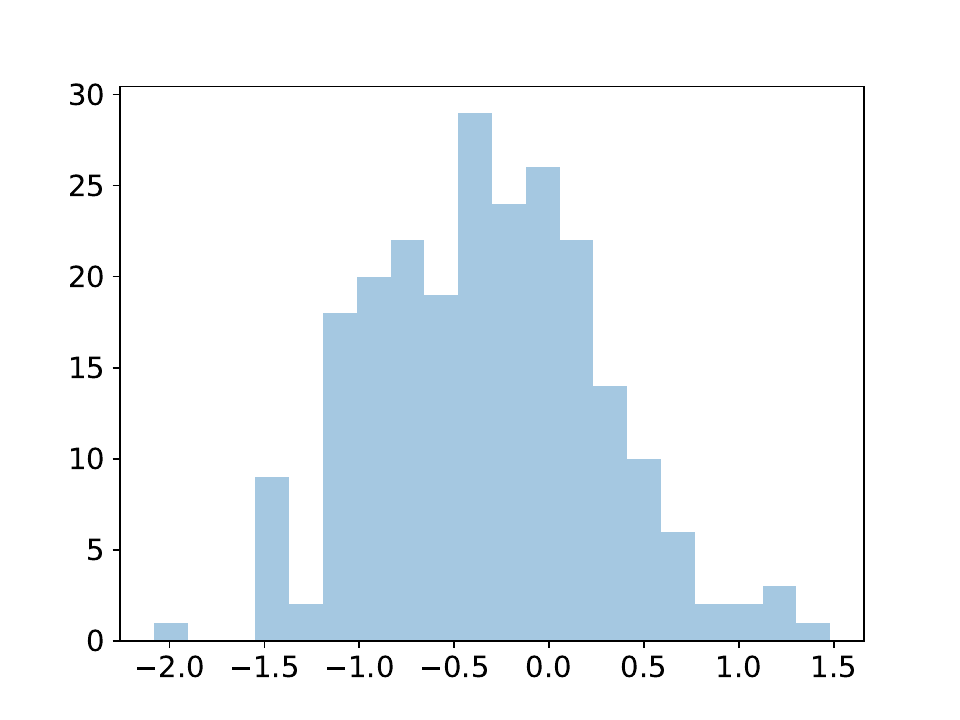}
    \subcaption{MMQA}
    \label{fig: MMQA_diff_hist}
  \end{minipage}
  \begin{minipage}{0.45\columnwidth}
    \centering
    \includegraphics[width=\columnwidth]{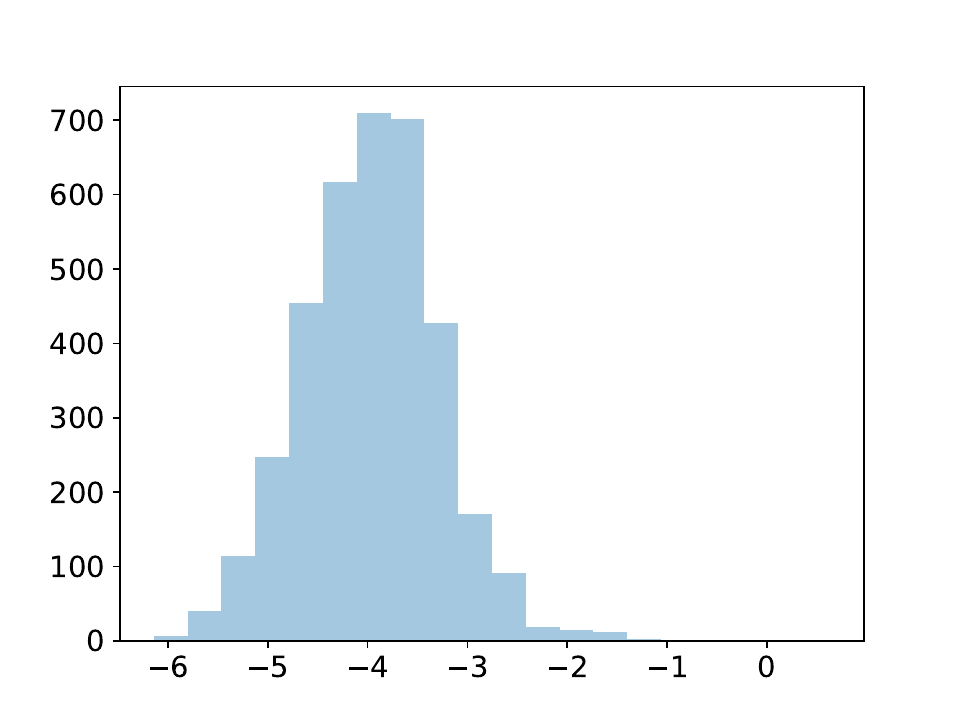}
    \subcaption{WebQA}
    \label{fig: WebQA_diff_hist}
  \end{minipage}
  \caption{Distributions of the difference between the average standardized similarity scores of image and text candidates across ImageQ queries in the training split of MMQA and WebQA, where the difference is computed as image minus text.}
  \label{fig: txt_img_diff_hist}
\end{figure}

To investigate how our method reduced the negative impact of the modality gap, we analyze the distribution of standardized similarity scores on ImageQ.
For each ImageQ, we compute the difference between the average standardized similarity scores for image and text candidates in the retrieval database~(image mean minus text mean).
The distributions on MMQA and WebQA are shown in Figure~\ref{fig: txt_img_diff_hist}, focusing on CLIP~(ViT-B/32) as a representative model that exhibits a clear modality gap.

In MMQA, the distribution is centered slightly below zero, indicating that text scores remain somewhat higher than image scores on average, even after the standardization.
In WebQA, the distribution is concentrated mostly on the negative side~(around $-4$), indicating that text candidates are consistently scored higher than images.
From these results, we confirm that our method does not fully eliminate the modality gap.

Nevertheless, retrieval performance improves significantly as shown in Section~\ref{sec: results}.
We attribute this to differences in the shape of the cosine similarity score distributions across modalities.
Table~\ref{table: skewness} shows the skewness values in the distributions of similarity scores.
CLIP-based models consistently produced more positively skewed similarity distributions for image candidates compared to text candidates.
This suggests that some images receive totally higher similarity scores than others in the image database.
Such outliers---which often include the correct images---were amplified by our method, allowing them to receive a higher standardized score than most text candidates.

We hypothesized that the skewness in the image similarity distribution stems from the training objective of CLIP-based models.
These models learn to align images with their paired text, but they are not explicitly trained to capture similarities between texts or between images themselves.
As a result, these models yield high similarities to a few image candidates, resulting in a long-tailed distribution.
This skewed distribution might align well with our standardization approach, as it amplifies the scores of outliers which often include relevant images.

\begin{table}[t]
  \centering
  \small
  \resizebox{\columnwidth}{!}{
    \begin{tabular}{lrrrr}
      \toprule
      \multirow{2}{*}{Model} & \multicolumn{2}{c}{MMQA} & \multicolumn{2}{c}{WebQA} \\
      \cmidrule(lr){2-3} \cmidrule(lr){4-5}
      & \multicolumn{1}{c}{Text} & \multicolumn{1}{c}{Image} & \multicolumn{1}{c}{Text} & \multicolumn{1}{c}{Image} \\
      \midrule
      CLIP~(ViT-B/32) & $-0.81$ & $0.21$ & $-1.05$ & $0.45$ \\
      CLIP~(ViT-L/14) & $-0.41$ & $0.31$ & $-0.51$ & $0.33$ \\
      Long-CLIP-B & $-1.40$ & $0.35$ & $-1.33$ & $0.59$ \\
      Long-CLIP-L & $-1.88$ & $0.47$ & $-1.30$ & $0.72$ \\
      BLIP & $0.32$ & $0.37$ & $0.59$ & $0.45$ \\
      Cohere Embed 3 & $0.26$ & $0.16$ & $0.54$ & $0.25$ \\
      E5-V & $0.81$ & $0.90$ & $0.91$ & $0.76$ \\
      \bottomrule
    \end{tabular}
  }
  \caption{Average skewnesses of cosine similarity distributions for ImageQ queries in the training split of MMQA and WebQA. Each skewness is computed between a query and all candidates in the text or image database, then averaged across all queries per modality.}
  \label{table: skewness}
\end{table}

\subsection{Modality Gap in VLMs}
We analyze the modality gap in VLMs by investigating both the structure of the embedding space and the distribution of similarity scores.

\begin{figure}[t]
  \centering
  \begin{minipage}{0.45\columnwidth}
    \centering
    \includegraphics[width=\columnwidth]{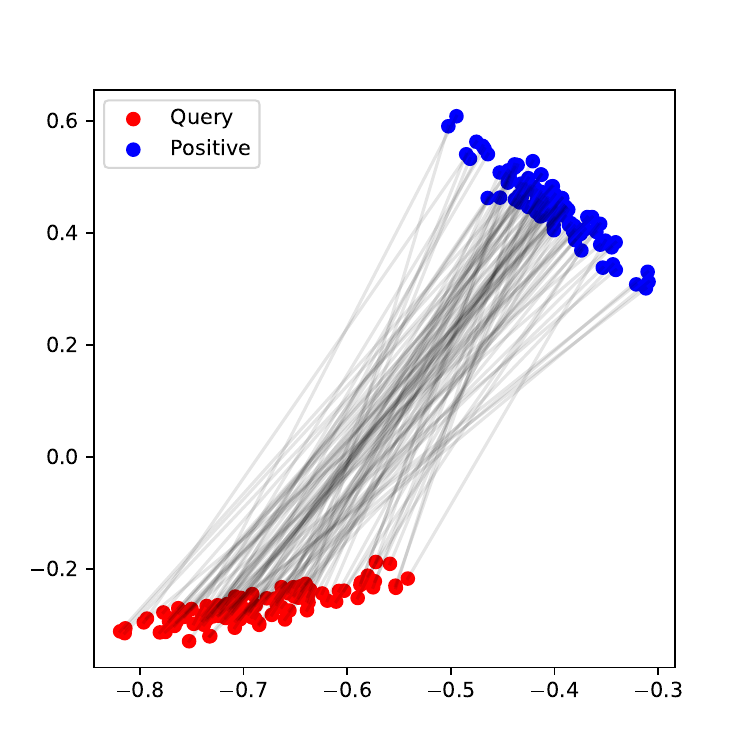}
    \subcaption{CLIP~(ViT-B/32)}
    \label{fig: svd_clip-base_img}
  \end{minipage}
  \begin{minipage}{0.45\columnwidth}
    \centering
    \includegraphics[width=\columnwidth]{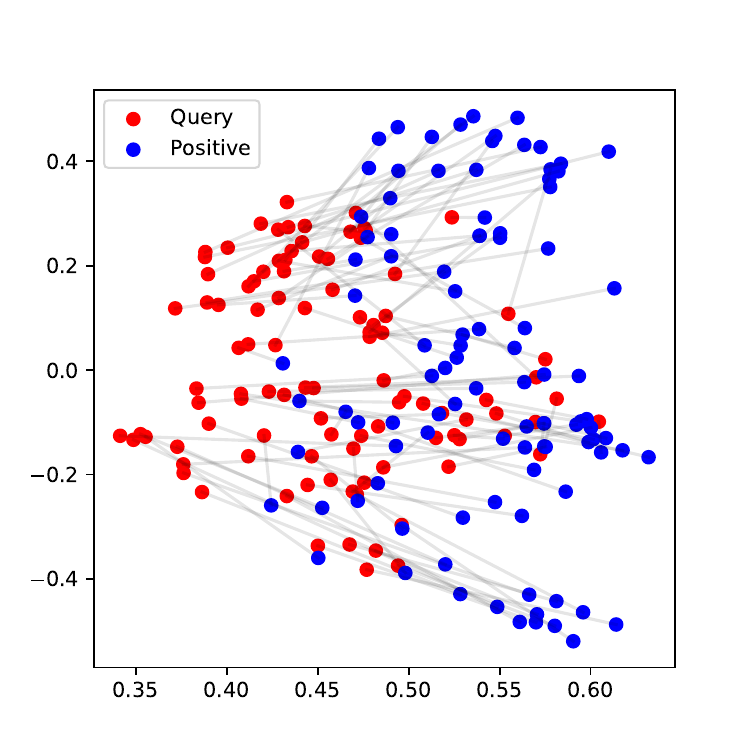}
    \subcaption{E5-V}
    \label{fig: svd_e5-v_img}
  \end{minipage}
  \caption{2D visualizations of the embeddings of ImageQ queries in the MMQA training split~(blue dots) and their corresponding images~(red dots) using SVD. Figures~\ref{fig: svd_clip-base_img} and \ref{fig: svd_e5-v_img} show the results of CLIP~(ViT-B/32) and E5-V, respectively.}
  \label{fig: svd_visualization}
\end{figure}

Following \citet{Liang2022}, we apply singular value decomposition~(SVD) to project the embeddings of ImageQ queries and their positive examples into a two-dimensional space for visualization.
Figure~\ref{fig: svd_visualization} shows the results for CLIP~(ViT-B/32) and E5-V.
The visualizations of other models and datasets are shown in Appendix~\ref{sec: svd_all}.
CLIP exhibits a clear separation between textual queries and positive image items in the embedding space. 
In contrast, E5-V shows a much smaller gap, suggesting that modality conversion reduces representational disparity between text and images.

\begin{figure}[t]
  \centering
  \begin{minipage}{0.45\columnwidth}
    \centering
    \includegraphics[width=\columnwidth]{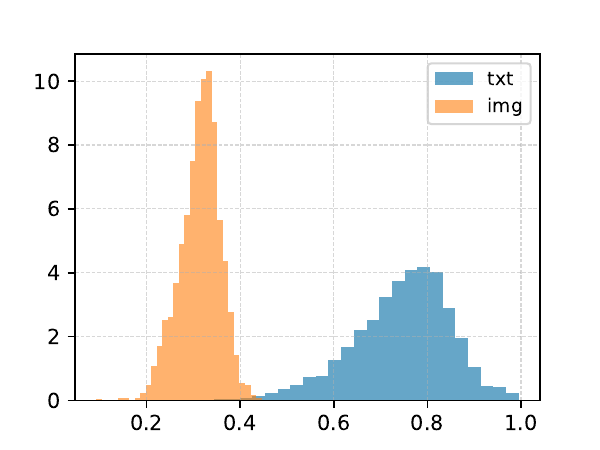}
    \subcaption{CLIP~(ViT-B/32)}
    \label{fig: cos_hist_clip-base_img}
  \end{minipage}
  \begin{minipage}{0.45\columnwidth}
    \centering
    \includegraphics[width=\columnwidth]{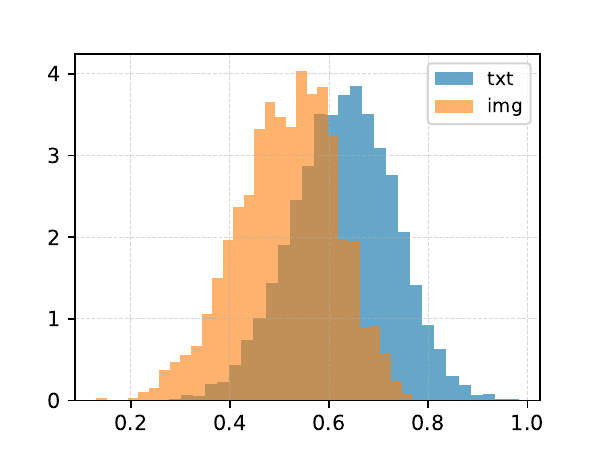}
    \subcaption{E5-V}
    \label{fig: cos_hist_e5-v_img}
  \end{minipage}
  \caption{Distributions of cosine similarity scores between a query and its corresponding positive example~(either text or image). The distributions are separated by the modality of the positive example. Figures~\ref{fig: cos_hist_clip-base_img} and \ref{fig: cos_hist_e5-v_img} show the results of CLIP~(ViT-B/32) and E5-V, respectively.}
  \label{fig: cos_hist}
\end{figure}

We then analyze the cosine similarity scores between queries in the training split of MMQA and their positive examples~(either text or image) for CLIP~(ViT-B/32) and E5-V.
Figure~\ref{fig: cos_hist} presents the distributions of these scores, separated by the modality of the positive examples.
The distributions of other models are shown in Appendix~\ref{sec: hist_all}.
As expected, CLIP assigns significantly higher similarities to text examples.
E5-V reduces this gap to some extent, but a consistent score difference remains: image positives still tend to receive lower similarity scores than text counterparts.

These results indicate that image captioning reduces modality differences, but does not fully avoid the gap of VLMs.
One possible reason is that converting images into textual representations leads to loss of visual information necessary for questions that are difficult to express in language, such as the spatial relationships between objects and the background color.
This missing information reduces similarities between queries and relevant candidates compared to text data.
Our method avoids this shortcoming.
By directly processing image features without converting them into text, our method outperformed E5-V in ImageQ.

\section{Conclusion}
We presented a method for improving multi-modal retrieval by bridging the modality gap without human-created data.
Our approach standardizes similarity scores in a modality-specific manner, making them more comparable across modalities.
Importantly, it does not require any labeled data or image captions, as it relies on pseudo-positive examples derived from unlabeled queries.
Through experiments on two multi-modal QA datasets and seven vision-language models, we demonstrated that our method consistently improves image retrieval performance, particularly in scenarios where existing models struggle due to the modality gap.
Furthermore, we showed that pseudo-positive examples are sufficient for estimating modality-specific statistics, achieving performance on par with manually labeled data.
Our findings highlight the importance of preserving modality-specific information and calibrating similarity scores, rather than relying solely on modality conversion.

\section*{Limitations}
Our method computes modality-specific similarity statistics from pre-collected datasets and uses them to standardize all similarity scores across modalities.
However, this approach assumes that similarity distributions remain stable over time.
In real-world systems, new data is constantly being added to databases.
Due to new content, these pre-computed statistics may become obsolete, leading to suboptimal standardization.
Future work should focus on developing mechanisms to dynamically update these statistics.

\bibliography{custom}

\appendix
\section{Computational Resources}
\label{subsec: resources}
We used two NVIDIA Quadro RTX 6000 GPUs for generating embeddings with E5-V, while only one GPU was used for all other pre-trained VLMs.
All retrieval and evaluation experiments were conducted using Faiss~\cite{faiss} on CPU only. 

\section{Model List}
\label{subsec: vlm_checkpoints}
We evaluated seven pre-trained VLMs in our experiments.  
Six of them are publicly available on Hugging Face and were accessed as downloadable checkpoints:
\begin{itemize}
  \item \url{https://huggingface.co/openai/clip-vit-base-patch32}
  \item \url{https://huggingface.co/openai/clip-vit-large-patch14}
  \item \url{https://huggingface.co/BeichenZhang/LongCLIP-B}
  \item \url{https://huggingface.co/BeichenZhang/LongCLIP-L}
  \item \url{https://huggingface.co/Salesforce/blip-itm-base-coco}
  \item \url{https://huggingface.co/royokong/e5-v}
\end{itemize}
We used the Cohere Embed 3 English model~(cohere.embed-english-v3) via Amazon Bedrock API in the us-west-2 region.

\section{Modality-Specific Mean and Variance}
\label{subsec: mean_variance}
Table~\ref{table: standardization_statistics} lists the modality-specific mean and standard deviation for similarity standardization that were used for standardization in our experiments.
\begin{table*}
  \centering
  \small
  \begin{tabular}{llrrrrrrrr}
    \toprule
    \multirow{2}{*}{Model} & \multirow{2}{*}{Method} & \multicolumn{4}{c}{MMQA} & \multicolumn{4}{c}{WebQA} \\
    & & \multicolumn{2}{c}{Text} & \multicolumn{2}{c}{Image} & \multicolumn{2}{c}{Text} & \multicolumn{2}{c}{Image} \\
    \cmidrule(lr){3-6} \cmidrule(lr){7-10}
    & & Mean & Std & Mean & Std & Mean & Std & Mean & Std \\
    \midrule
    \multirow{2}{*}{CLIP~(ViT-B/32)} & Std & $0.744$ & $0.105$ & $0.314$ & $0.043$ & $0.789$ & $0.093$ & $0.304$ & $0.035$ \\
                                     & Ours & $0.841$ & $0.058$ & $0.315$ & $0.023$ & $0.833$ & $0.063$ & $0.335$ & $0.019$ \\
    \cmidrule{1-10}
    \multirow{2}{*}{CLIP~(ViT-L/14)} & Std & $0.642$ & $0.136$ & $0.280$ & $0.049$ & $0.700$ & $0.122$ & $0.269$ & $0.040$ \\
                                     & Ours & $0.755$ & $0.088$ & $0.271$ & $0.029$ & $0.749$ & $0.093$ & $0.297$ & $0.023$ \\
    \cmidrule{1-10}
    \multirow{2}{*}{Long-CLIP-B} & Std & $0.879$ & $0.050$ & $0.315$ & $0.031$ & $0.895$ & $0.040$ & $0.307$ & $0.024$ \\
                                 & Ours & $0.898$ & $0.043$ & $0.311$ & $0.017$ & $0.901$ & $0.037$ & $0.324$ & $0.016$ \\
    \cmidrule{1-10}
    \multirow{2}{*}{Long-CLIP-L} & Std & $0.828$ & $0.068$ & $0.279$ & $0.048$ & $0.856$ & $0.057$ & $0.258$ & $0.037$ \\
                                 & Ours & $0.845$ & $0.073$ & $0.264$ & $0.029$ & $0.860$ & $0.059$ & $0.277$ & $0.026$ \\
    \cmidrule{1-10}
    \multirow{2}{*}{BLIP} & Std & $0.700$ & $0.116$ & $0.438$ & $0.072$ & $0.724$ & $0.100$ & $0.418$ & $0.059$ \\
                          & Ours & $0.791$ & $0.070$ & $0.460$ & $0.038$ & $0.806$ & $0.058$ & $0.489$ & $0.034$ \\
    \cmidrule{1-10}
    \multirow{2}{*}{Cohere Embed 3} & Std & $0.629$ & $0.121$ & $0.508$ & $0.082$ & $0.581$ & $0.114$ & $0.490$ & $0.066$ \\
                                    & Ours & $0.660$ & $0.105$ & $0.512$ & $0.047$ & $0.615$ & $0.082$ & $0.541$ & $0.044$ \\
    \cmidrule{1-10}
    \multirow{2}{*}{E5-V} & Std & $0.628$ & $0.102$ & $0.514$ & $0.099$ & $0.635$ & $0.105$ & $0.467$ & $0.084$ \\
                          & Ours & $0.649$ & $0.095$ & $0.469$ & $0.085$ & $0.640$ & $0.093$ & $0.534$ & $0.073$ \\
    \bottomrule
  \end{tabular}
  \caption{Modality-specific mean and standard deviation used for standardization during evaluation on MMQA and WebQA datasets. Values are computed separately for text and image modalities, either from labeled or pseudo pairs.}
  \label{table: standardization_statistics}
\end{table*}

\section{2D Visualizations of Embeddings}
\label{sec: svd_all}
\begin{figure*}[t]
  \centering
  \begin{subfigure}{0.23\textwidth}
    \centering
    \includegraphics[width=\textwidth]{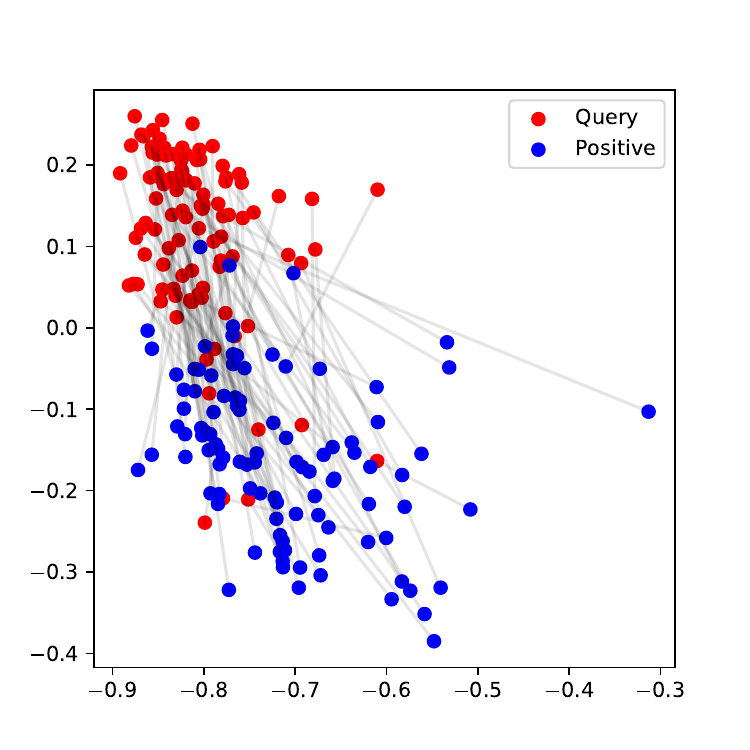}
    \caption{MMQA TextQ}
    \label{fig: MMQA_clip_base_txt}
  \end{subfigure}
  \begin{subfigure}{0.23\textwidth}
    \centering
    \includegraphics[width=\textwidth]{figures/svd/MMQA_clip-base_img_svd.pdf}
    \caption{MMQA ImageQ}
    \label{fig: MMQA_clip_base_img}
  \end{subfigure}
  \begin{subfigure}{0.23\textwidth}
    \centering
    \includegraphics[width=\textwidth]{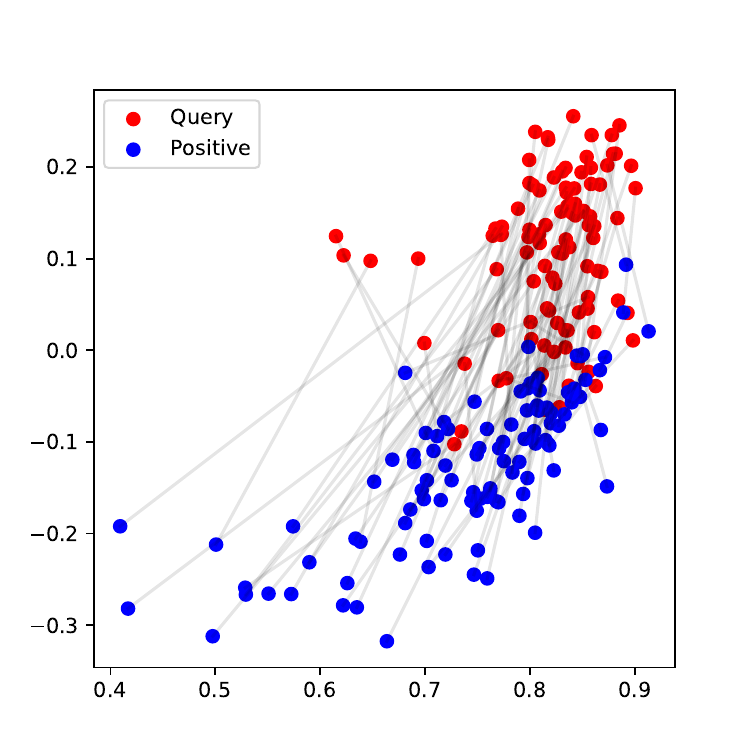}
    \caption{WebQA TextQ}
    \label{fig: WebQA_clip_base_txt}
  \end{subfigure}
  \begin{subfigure}{0.23\textwidth}
    \centering
    \includegraphics[width=\textwidth]{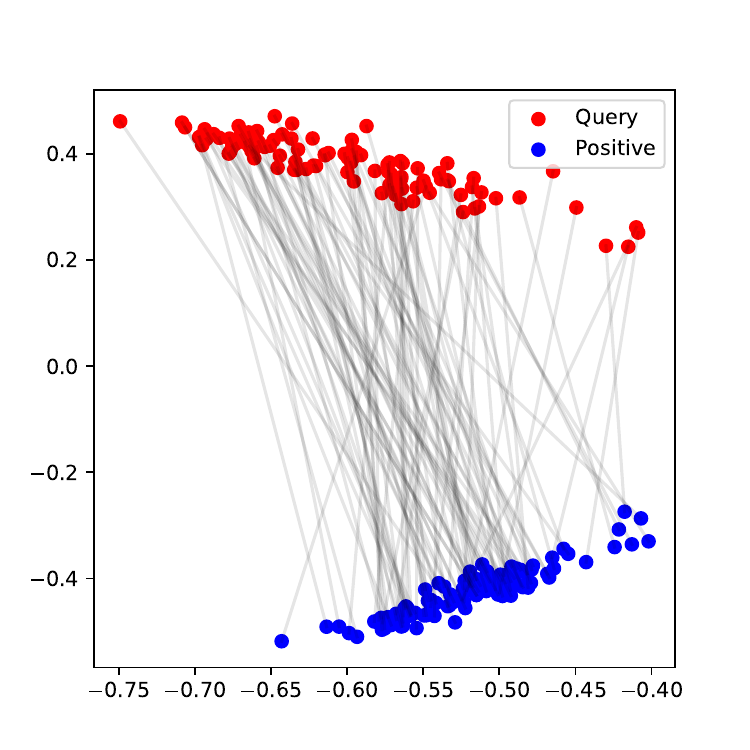}
    \caption{WebQA ImageQ}
    \label{fig: WebQA_clip_base_img}
  \end{subfigure}
  \caption{2D visualizations of embeddings from CLIP~(ViT-B/32).}
  \label{fig: svd_CLIP_base}
\end{figure*}

\begin{figure*}[t]
  \centering
  \begin{subfigure}{0.23\textwidth}
    \centering
    \includegraphics[width=\textwidth]{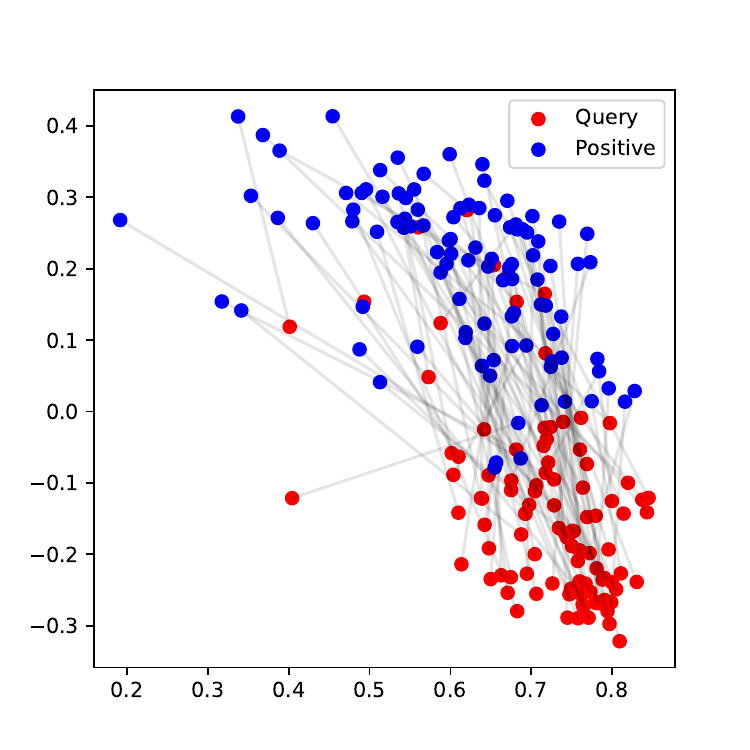}
    \caption{MMQA TextQ}
    \label{fig: MMQA_clip_large_txt}
  \end{subfigure}
  \begin{subfigure}{0.23\textwidth}
    \centering
    \includegraphics[width=\textwidth]{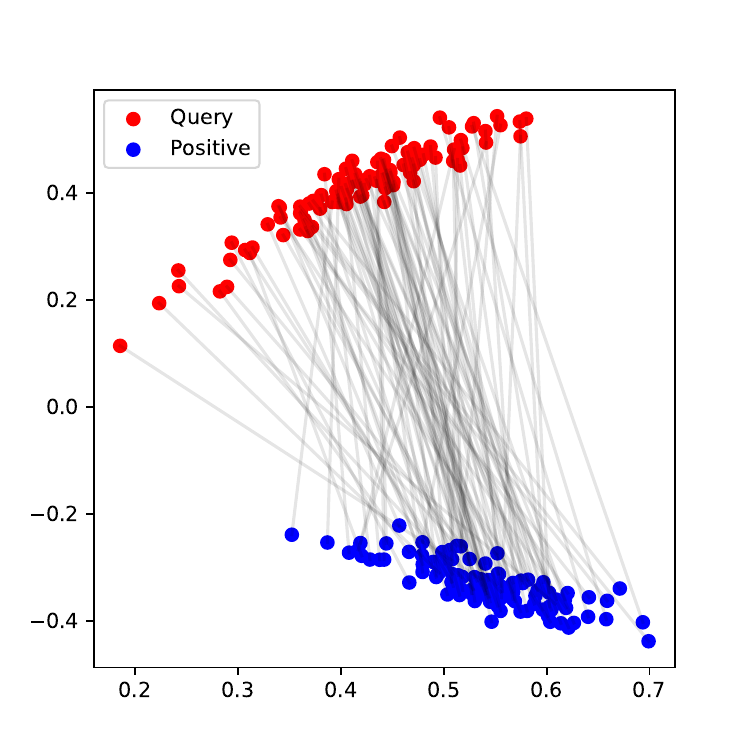}
    \caption{MMQA ImageQ}
    \label{fig: MMQA_clip_large_img}
  \end{subfigure}
  \begin{subfigure}{0.23\textwidth}
    \centering
    \includegraphics[width=\textwidth]{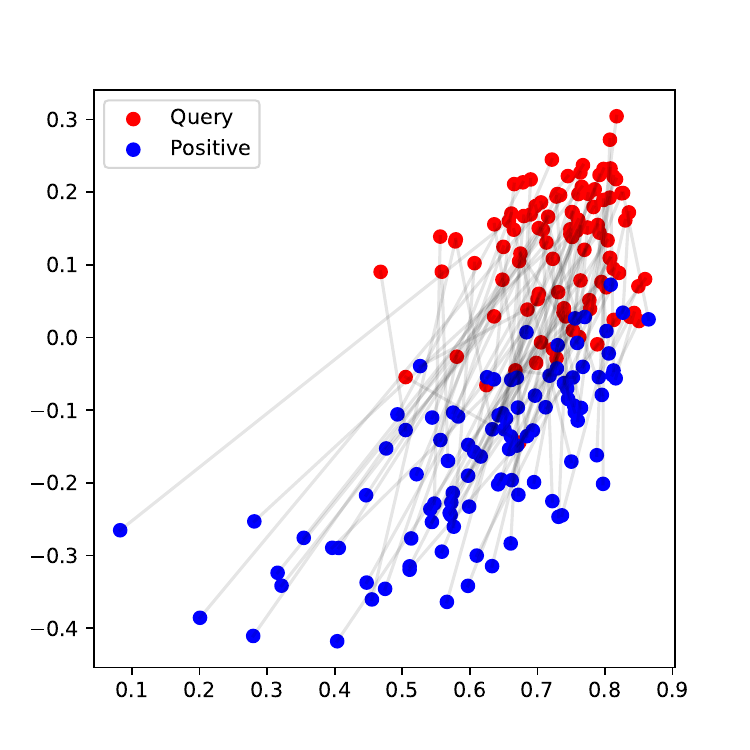}
    \caption{WebQA TextQ}
    \label{fig: WebQA_clip_large_txt}
  \end{subfigure}
  \begin{subfigure}{0.23\textwidth}
    \centering
    \includegraphics[width=\textwidth]{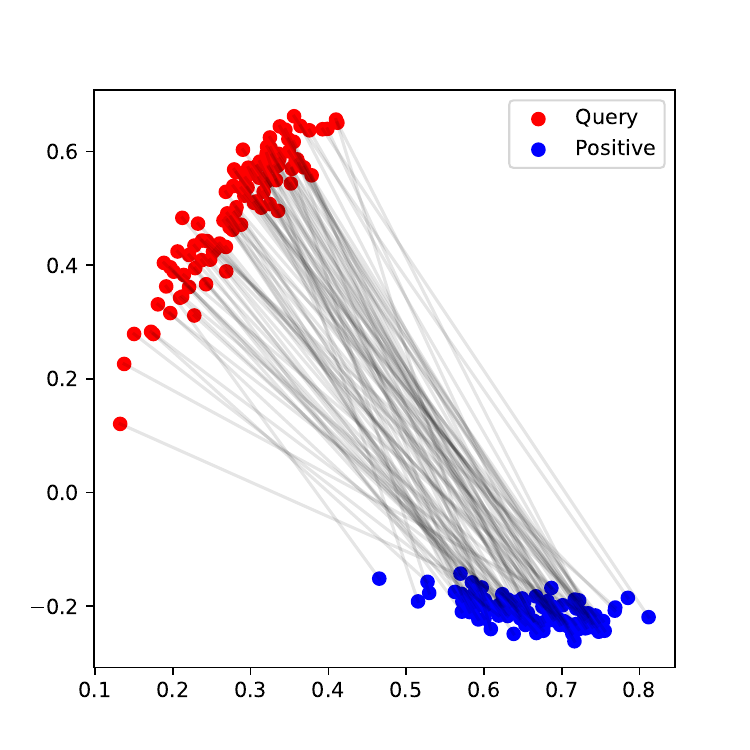}
    \caption{WebQA ImageQ}
    \label{fig: WebQA_clip_large_img}
  \end{subfigure}
  \caption{2D visualizations of embeddings from CLIP~(ViT-L/14).}
  \label{fig: svd_CLIP_large}
\end{figure*}

\begin{figure*}[t]
  \centering
  \begin{subfigure}{0.23\textwidth}
    \centering
    \includegraphics[width=\textwidth]{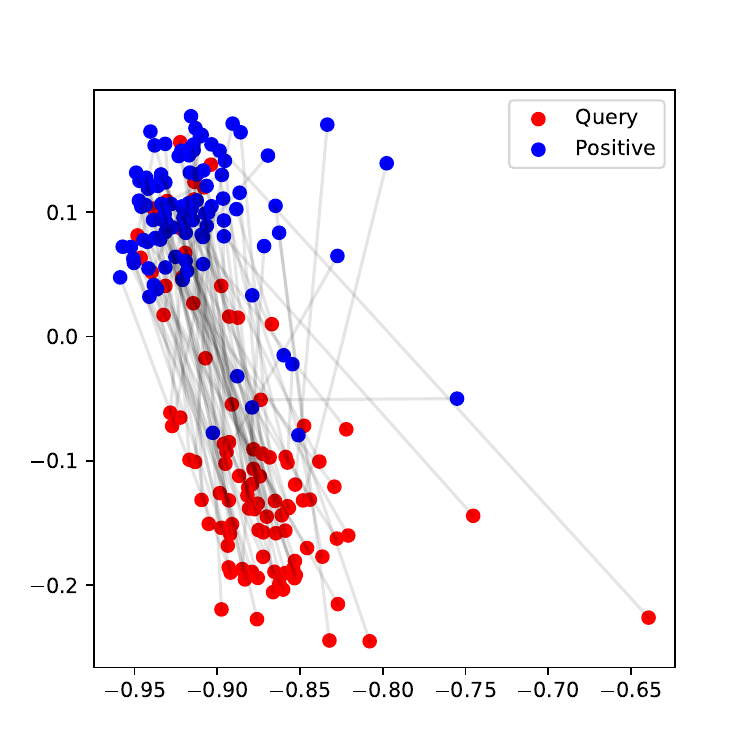}
    \caption{MMQA TextQ}
    \label{fig: MMQA_long_clip_base_txt}
  \end{subfigure}
  \begin{subfigure}{0.23\textwidth}
    \centering
    \includegraphics[width=\textwidth]{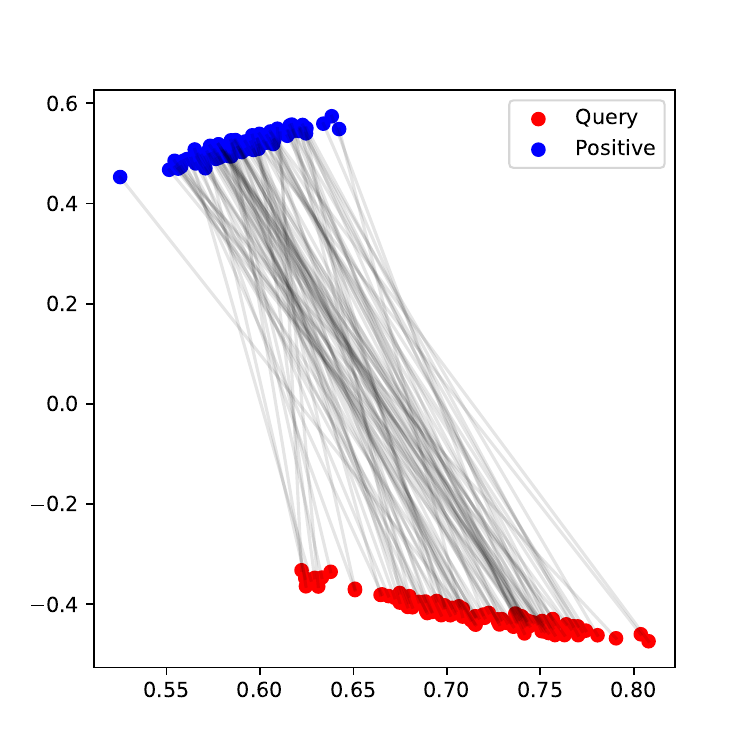}
    \caption{MMQA ImageQ}
    \label{fig: MMQA_long_clip_base_img}
  \end{subfigure}
  \begin{subfigure}{0.23\textwidth}
    \centering
    \includegraphics[width=\textwidth]{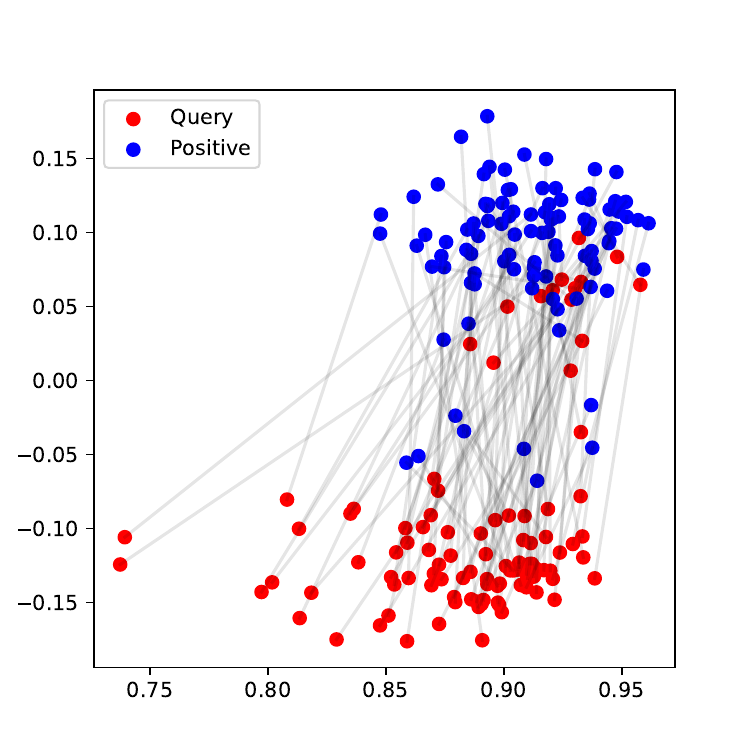}
    \caption{WebQA TextQ}
    \label{fig: WebQA_long_clip_base_txt}
  \end{subfigure}
  \begin{subfigure}{0.23\textwidth}
    \centering
    \includegraphics[width=\textwidth]{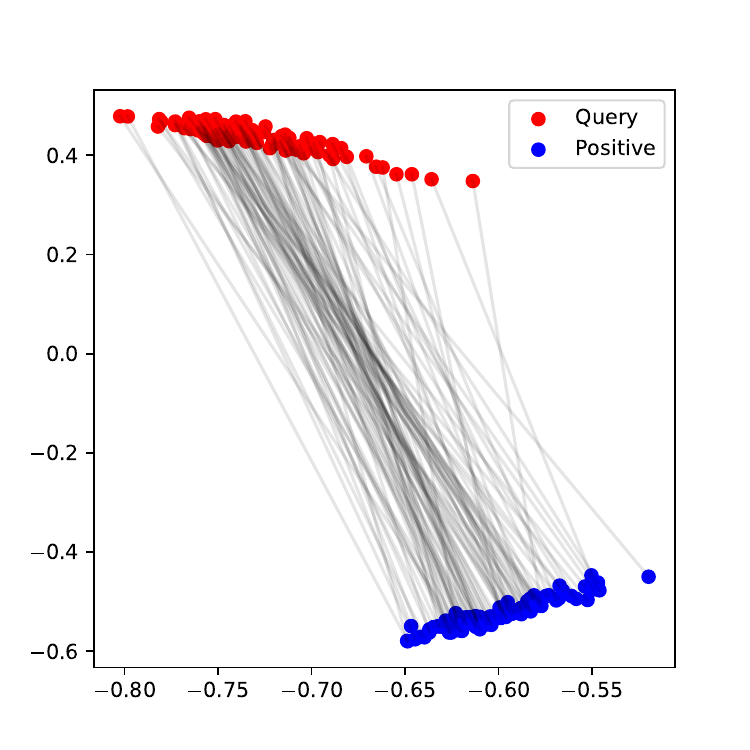}
    \caption{WebQA ImageQ}
    \label{fig: WebQA_long_clip_base_img}
  \end{subfigure}
  \caption{2D visualizations of embeddings from Long-CLIP-B.}
  \label{fig: svd_Long_CLIP_B}
\end{figure*}

\begin{figure*}[t]
  \centering
  \begin{subfigure}{0.23\textwidth}
    \centering
    \includegraphics[width=\textwidth]{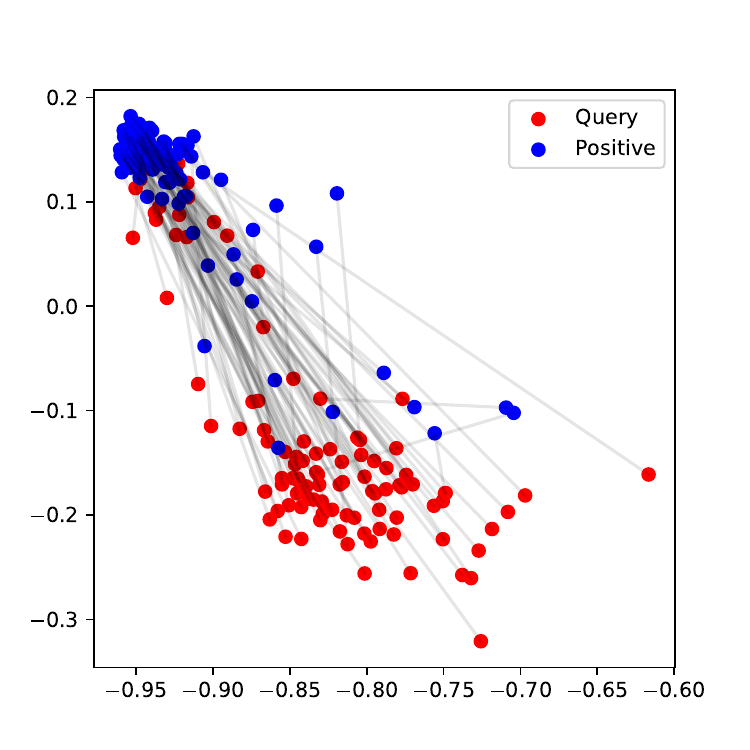}
    \caption{MMQA TextQ}
    \label{fig: MMQA_long_clip_large_txt}
  \end{subfigure}
  \begin{subfigure}{0.23\textwidth}
    \centering
    \includegraphics[width=\textwidth]{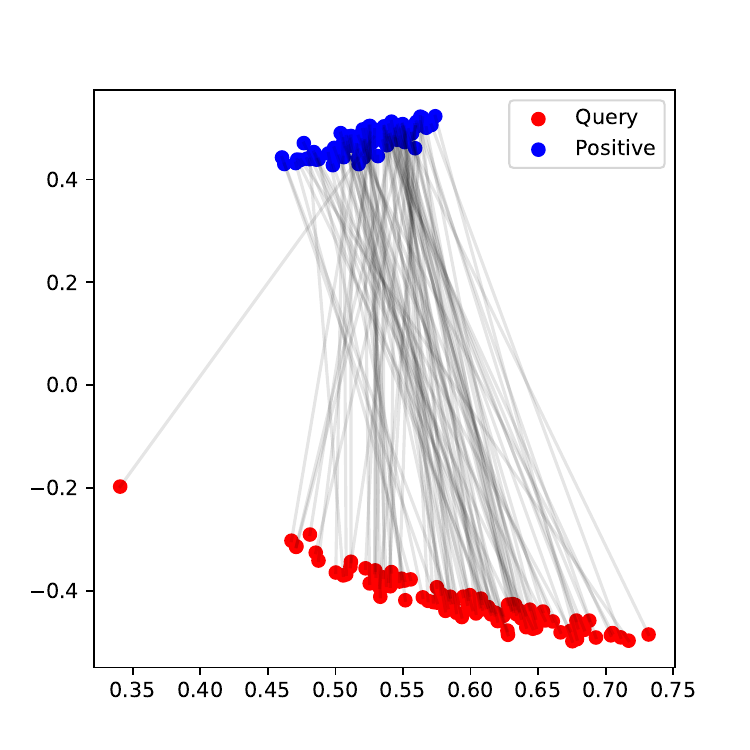}
    \caption{MMQA ImageQ}
    \label{fig: MMQA_long_clip_large_img}
  \end{subfigure}
  \begin{subfigure}{0.23\textwidth}
    \centering
    \includegraphics[width=\textwidth]{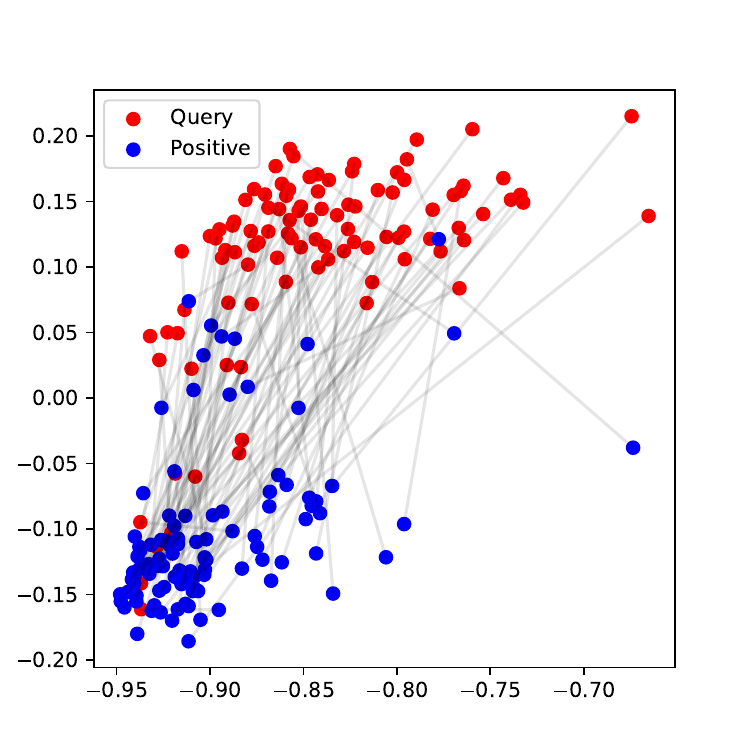}
    \caption{WebQA TextQ}
    \label{fig: WebQA_long_clip_large_txt}
  \end{subfigure}
  \begin{subfigure}{0.23\textwidth}
    \centering
    \includegraphics[width=\textwidth]{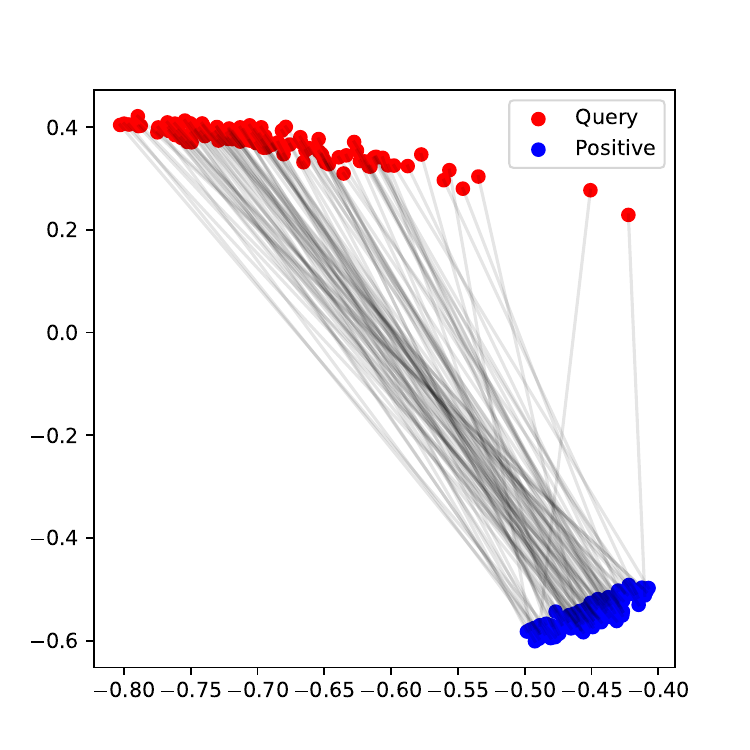}
    \caption{WebQA ImageQ}
    \label{fig: WebQA_long_clip_large_img}
  \end{subfigure}
  \caption{2D visualizations of embeddings from Long-CLIP-L.}
  \label{fig: svd_Long_CLIP_L}
\end{figure*}

\begin{figure*}[t]
  \centering
  \begin{subfigure}{0.23\textwidth}
    \centering
    \includegraphics[width=\textwidth]{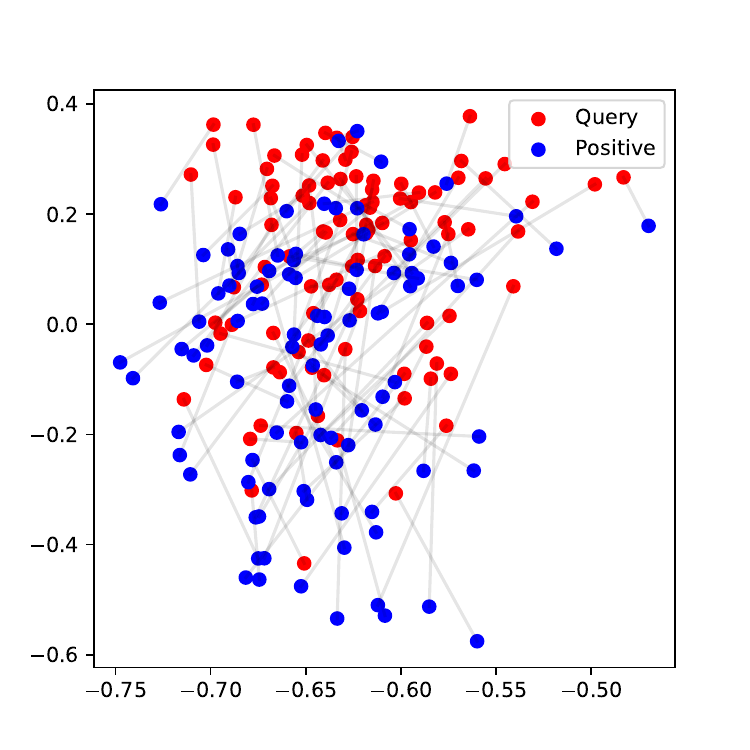}
    \caption{MMQA TextQ}
    \label{fig: MMQA_blip_txt}
  \end{subfigure}
  \begin{subfigure}{0.23\textwidth}
    \centering
    \includegraphics[width=\textwidth]{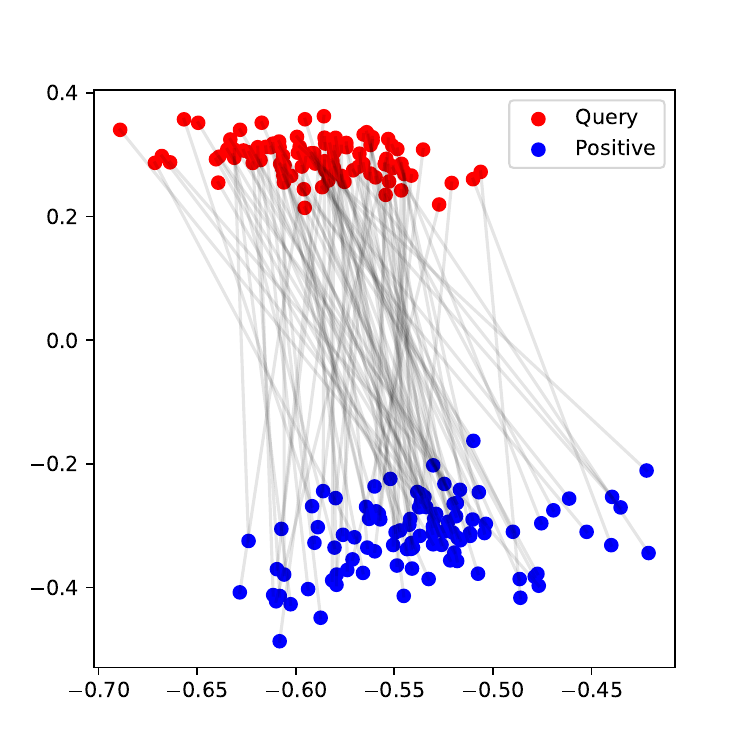}
    \caption{MMQA ImageQ}
    \label{fig: MMQA_blip_img}
  \end{subfigure}
  \begin{subfigure}{0.23\textwidth}
    \centering
    \includegraphics[width=\textwidth]{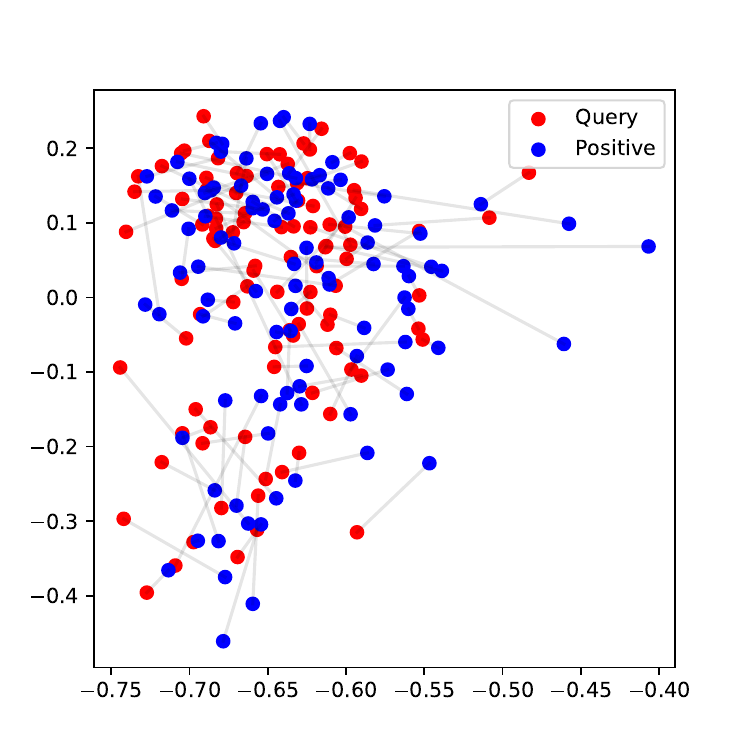}
    \caption{WebQA TextQ}
    \label{fig: WebQA_blip_txt}
  \end{subfigure}
  \begin{subfigure}{0.23\textwidth}
    \centering
    \includegraphics[width=\textwidth]{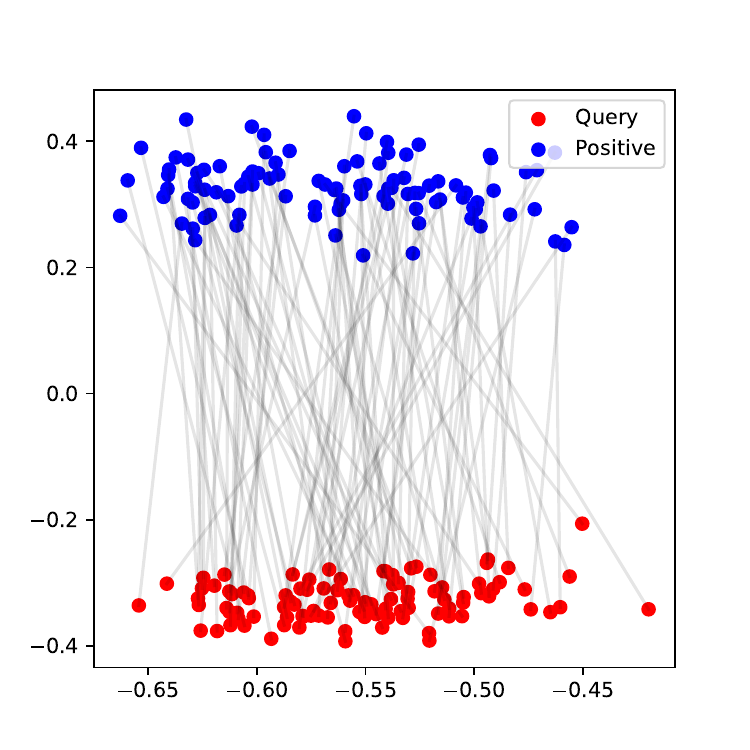}
    \caption{WebQA ImageQ}
    \label{fig: WebQA_blip_img}
  \end{subfigure}
  \caption{2D visualizations of embeddings from BLIP.}
  \label{fig: svd_BLIP}
\end{figure*}

\begin{figure*}[t]
  \centering
  \begin{subfigure}{0.23\textwidth}
    \centering
    \includegraphics[width=\textwidth]{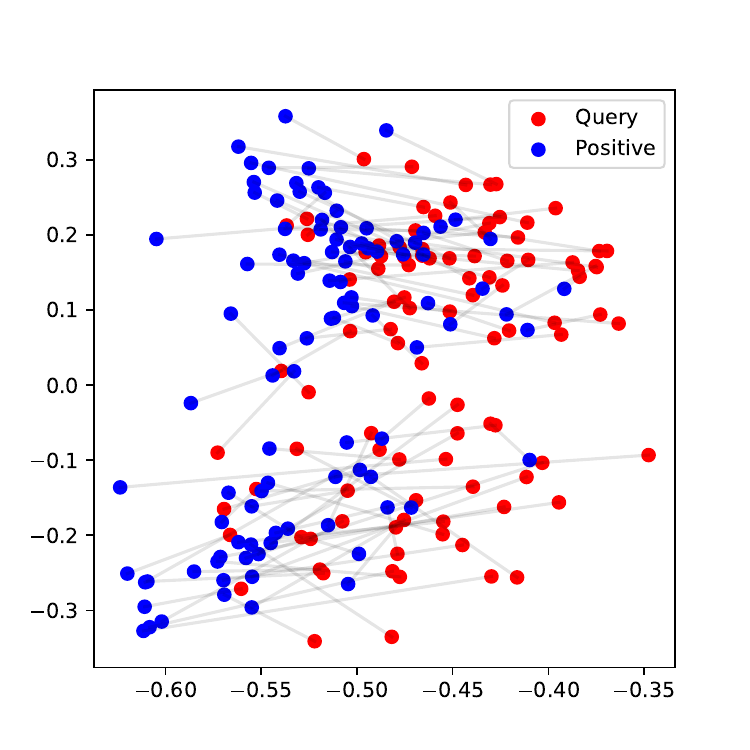}
    \caption{MMQA TextQ}
    \label{fig: MMQA_e5_txt}
  \end{subfigure}
  \begin{subfigure}{0.23\textwidth}
    \centering
    \includegraphics[width=\textwidth]{figures/svd/MMQA_e5-v_img_svd.pdf}
    \caption{MMQA ImageQ}
    \label{fig: MMQA_e5_img}
  \end{subfigure}
  \begin{subfigure}{0.23\textwidth}
    \centering
    \includegraphics[width=\textwidth]{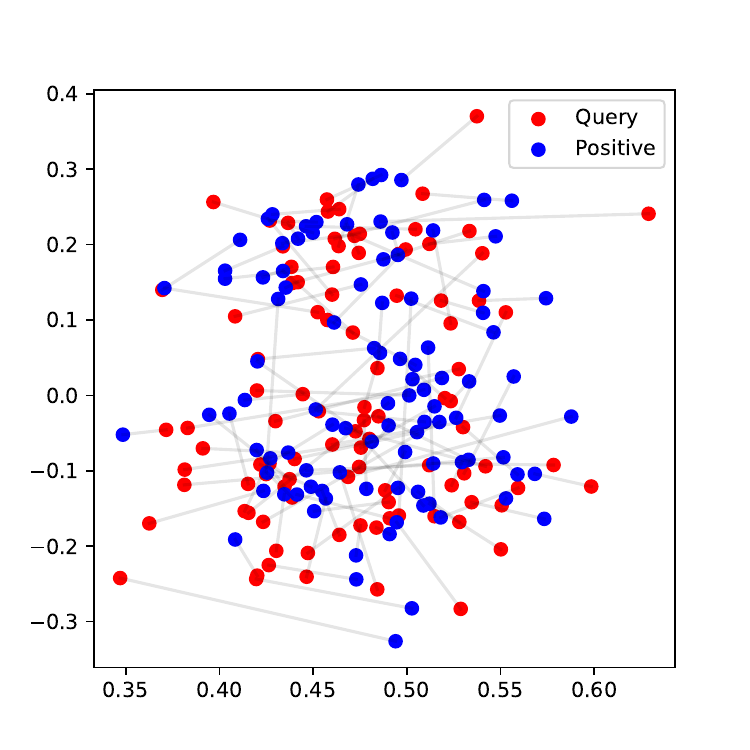}
    \caption{WebQA TextQ}
    \label{fig: WebQA_e5_txt}
  \end{subfigure}
  \begin{subfigure}{0.23\textwidth}
    \centering
    \includegraphics[width=\textwidth]{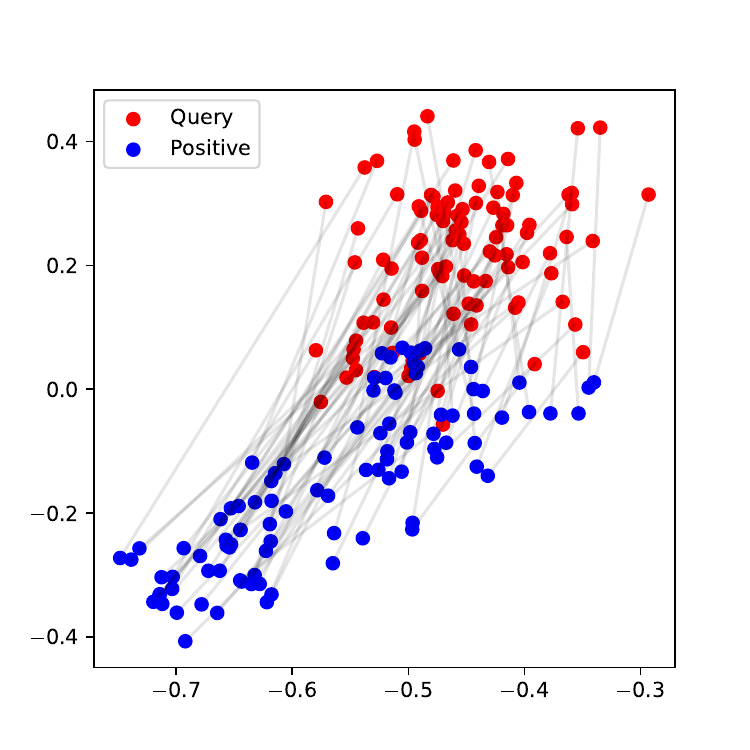}
    \caption{WebQA ImageQ}
    \label{fig: WebQA_e5_img}
  \end{subfigure}
  \caption{2D visualizations of embeddings from E5-V.}
  \label{fig: svd_E5_V}
\end{figure*}

\begin{figure*}[t]
  \centering
  \begin{subfigure}{0.23\textwidth}
    \centering
    \includegraphics[width=\textwidth]{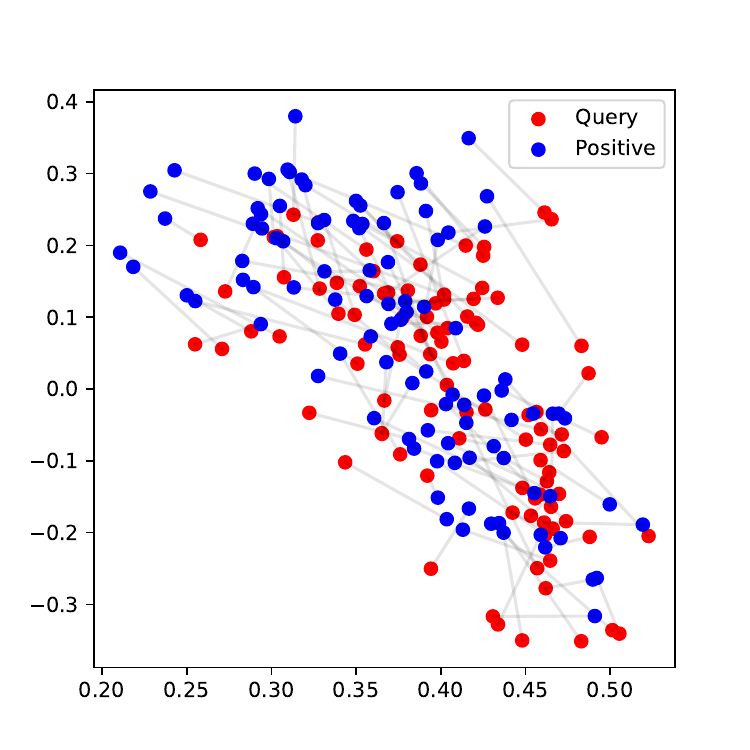}
    \caption{MMQA TextQ}
    \label{fig: MMQA_cohere_english_txt}
  \end{subfigure}
  \begin{subfigure}{0.23\textwidth}
    \centering
    \includegraphics[width=\textwidth]{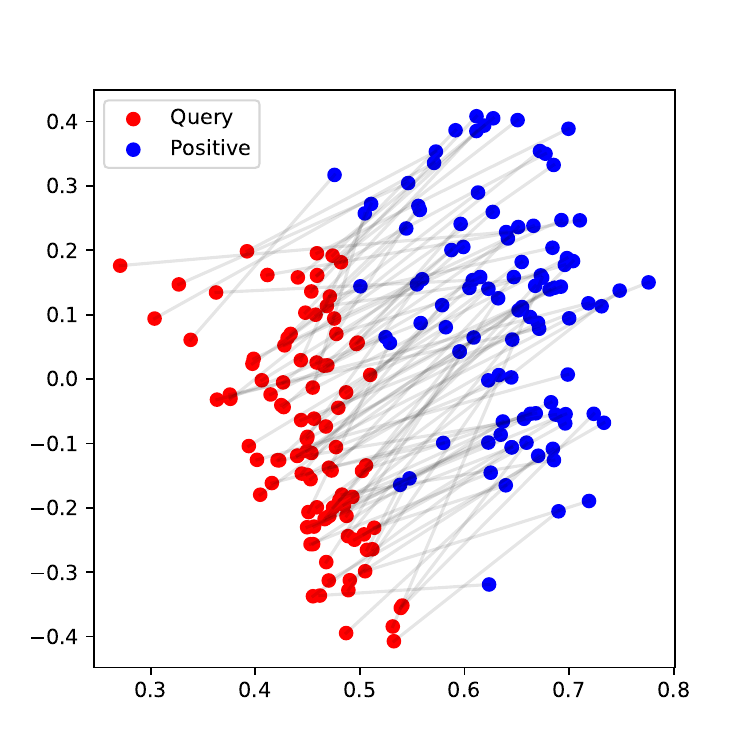}
    \caption{MMQA ImageQ}
    \label{fig: MMQA_cohere_english_img}
  \end{subfigure}
  \begin{subfigure}{0.23\textwidth}
    \centering
    \includegraphics[width=\textwidth]{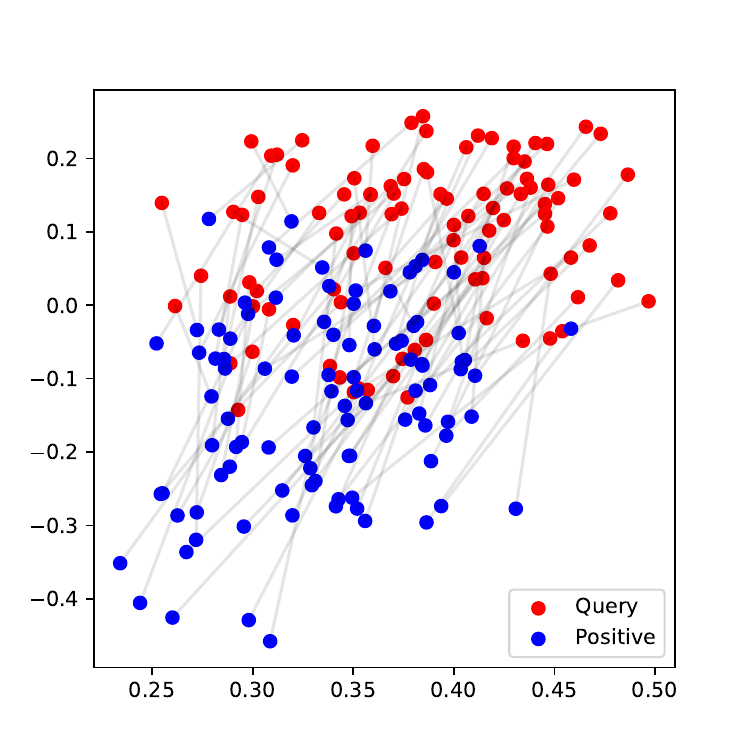}
    \caption{WebQA TextQ}
    \label{fig: WebQA_cohere_english_txt}
  \end{subfigure}
  \begin{subfigure}{0.23\textwidth}
    \centering
    \includegraphics[width=\textwidth]{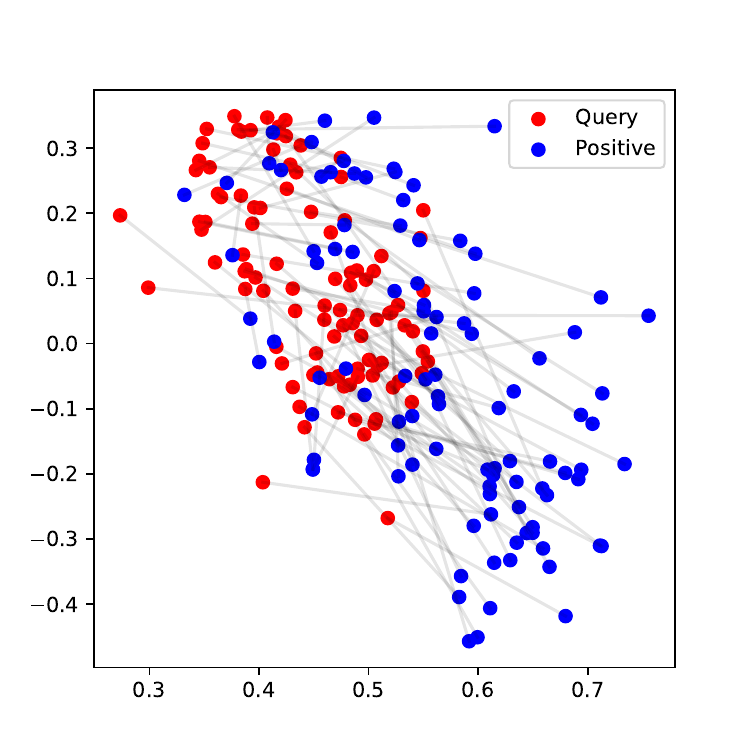}
    \caption{WebQA ImageQ}
    \label{fig: WebQA_cohere_english_img}
  \end{subfigure}
  \caption{2D visualizations of embeddings from Cohere Embed 3 English.}
  \label{fig: svd_Cohere_English}
\end{figure*}

Figures~\ref{fig: svd_CLIP_base}--\ref{fig: svd_Cohere_English} illustrate 2D visualizations of embeddings of textual queries~(from the training sets of MMQA and WebQA) and their positive examples using singular value decomposition\footnote{Our visualization code is adapted from \url{https://github.com/Weixin-Liang/Modality-Gap/blob/main/Figure_1_Modality_Gap/visualize.ipynb}}.

\section{Distributions of Cosine Similarity Scores between Positive Pairs across Modalities}
\label{sec: hist_all}
\begin{figure*}[t]
  \centering
  \begin{minipage}[t]{0.48\textwidth}
    \centering
    \begin{subfigure}[b]{0.48\textwidth}
      \includegraphics[width=\linewidth]{figures/txt_vs_img_hists/MMQA_cos_hist_clip-base.pdf}
      \caption{MMQA}
    \end{subfigure}
    \begin{subfigure}[b]{0.48\textwidth}
      \includegraphics[width=\linewidth]{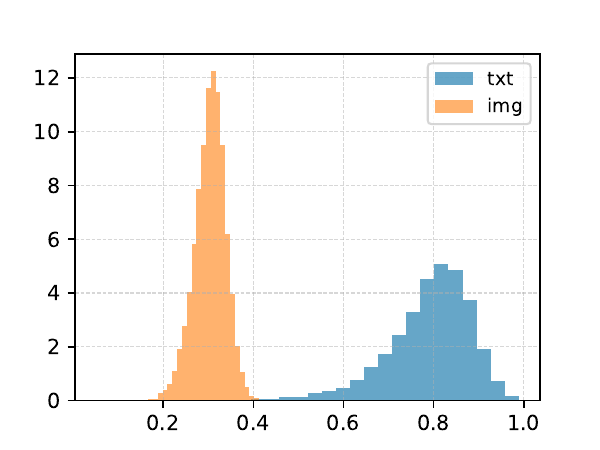}
      \caption{WebQA}
    \end{subfigure}
    \caption*{(A) CLIP~(ViT-B/32)}
  \end{minipage}
  \hfill
  \begin{minipage}[t]{0.48\textwidth}
    \centering
    \begin{subfigure}[b]{0.48\textwidth}
      \includegraphics[width=\linewidth]{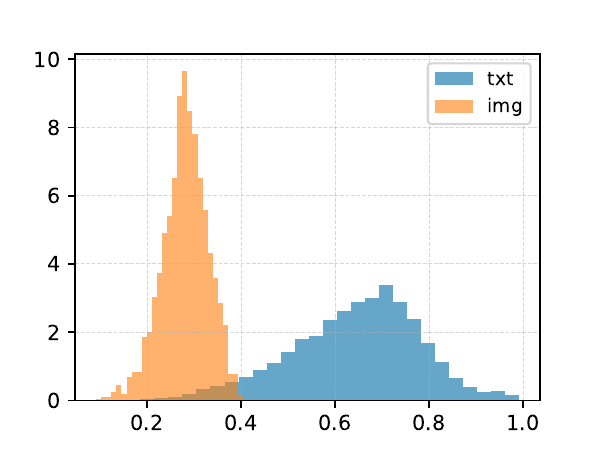}
      \caption{MMQA}
    \end{subfigure}
    \begin{subfigure}[b]{0.48\textwidth}
      \includegraphics[width=\linewidth]{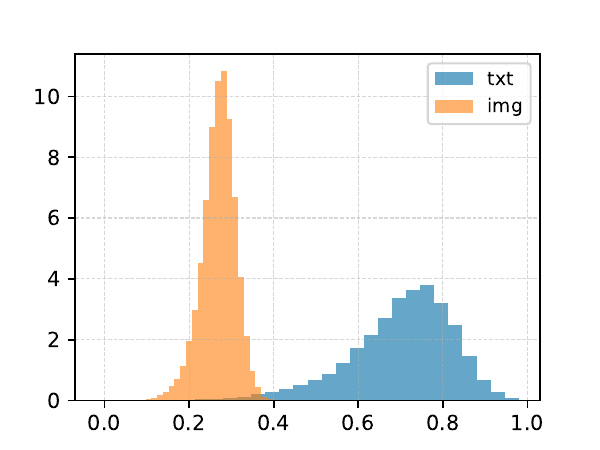}
      \caption{WebQA}
    \end{subfigure}
    \caption*{(B) CLIP~(ViT-L/14)}
  \end{minipage}

  \vspace{1em}

  \begin{minipage}[t]{0.48\textwidth}
    \centering
    \begin{subfigure}[b]{0.48\textwidth}
      \includegraphics[width=\linewidth]{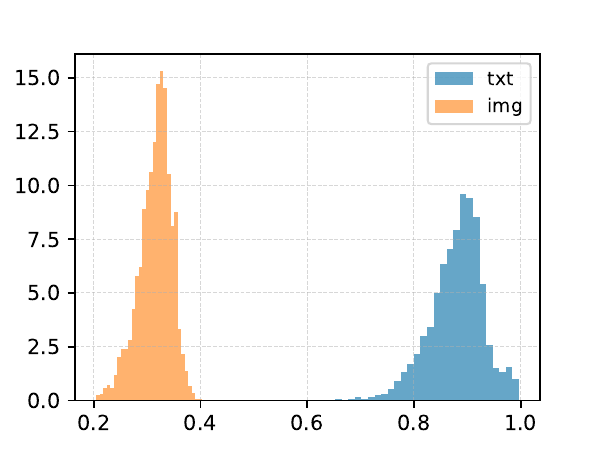}
      \caption{MMQA}
    \end{subfigure}
    \begin{subfigure}[b]{0.48\textwidth}
      \includegraphics[width=\linewidth]{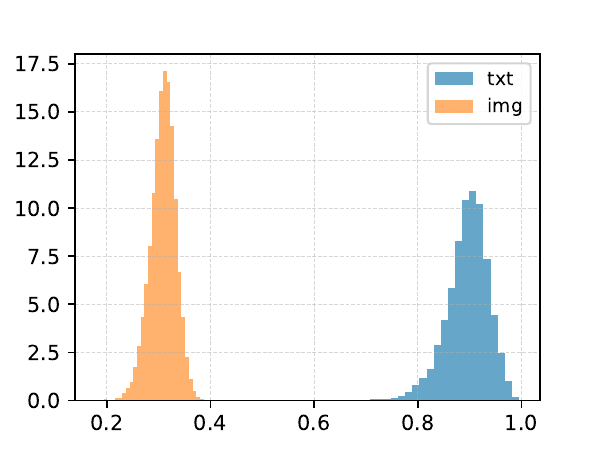}
      \caption{WebQA}
    \end{subfigure}
    \caption*{(C) Long-CLIP-B}
  \end{minipage}
  \hfill
  \begin{minipage}[t]{0.48\textwidth}
    \centering
    \begin{subfigure}[b]{0.48\textwidth}
      \includegraphics[width=\linewidth]{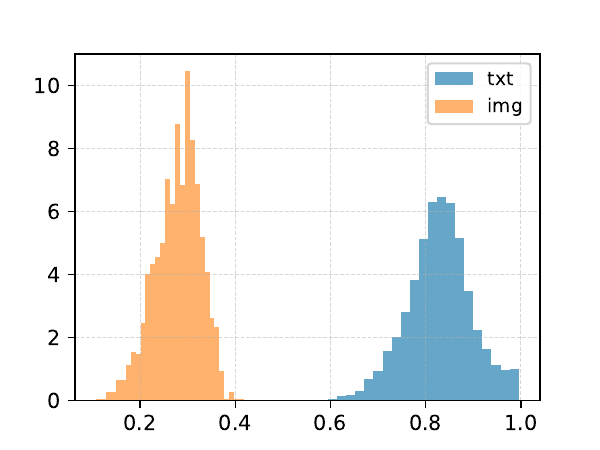}
      \caption{MMQA}
    \end{subfigure}
    \begin{subfigure}[b]{0.48\textwidth}
      \includegraphics[width=\linewidth]{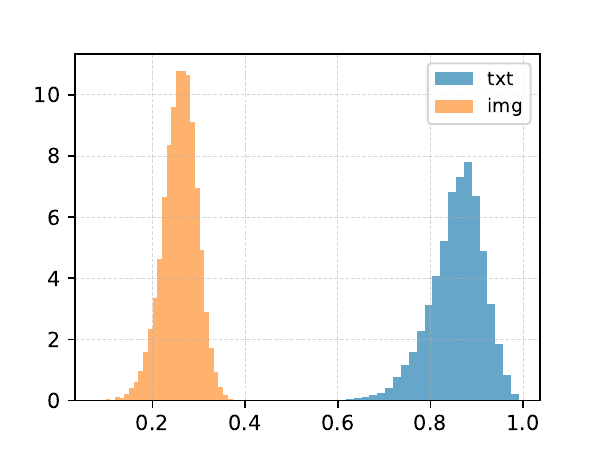}
      \caption{WebQA}
    \end{subfigure}
    \caption*{(D) Long-CLIP-L}
  \end{minipage}

  \vspace{1em}

  \begin{minipage}[t]{0.48\textwidth}
    \centering
    \begin{subfigure}[b]{0.48\textwidth}
      \includegraphics[width=\linewidth]{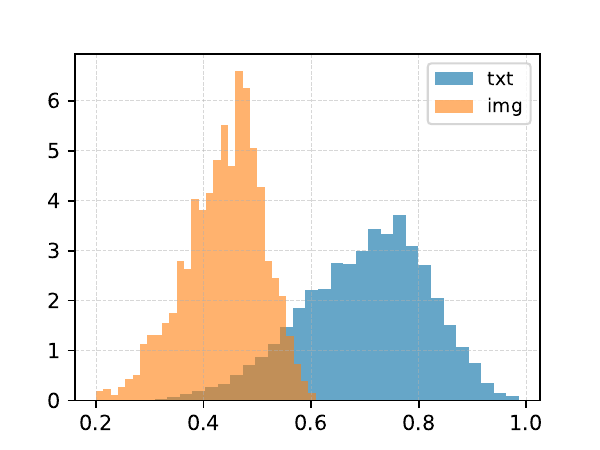}
      \caption{MMQA}
    \end{subfigure}
    \begin{subfigure}[b]{0.48\textwidth}
      \includegraphics[width=\linewidth]{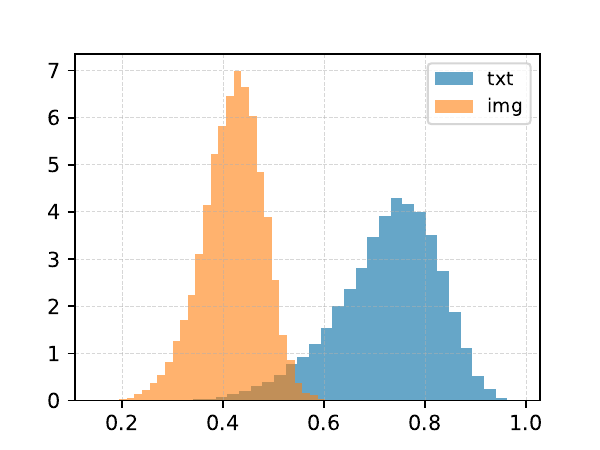}
      \caption{WebQA}
    \end{subfigure}
    \caption*{(E) BLIP}
  \end{minipage}
  \hfill
  \begin{minipage}[t]{0.48\textwidth}
    \centering
    \begin{subfigure}[b]{0.48\textwidth}
      \includegraphics[width=\linewidth]{figures/txt_vs_img_hists/MMQA_cos_hist_e5-v.pdf}
      \caption{MMQA}
    \end{subfigure}
    \begin{subfigure}[b]{0.48\textwidth}
      \includegraphics[width=\linewidth]{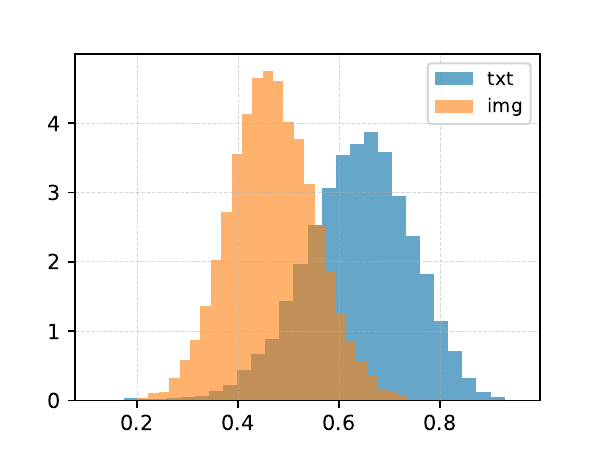}
      \caption{WebQA}
    \end{subfigure}
    \caption*{(F) E5-V}
  \end{minipage}

  \vspace{1em}

  \begin{minipage}[t]{0.48\textwidth}
    \centering
    \begin{subfigure}[b]{0.48\textwidth}
      \includegraphics[width=\linewidth]{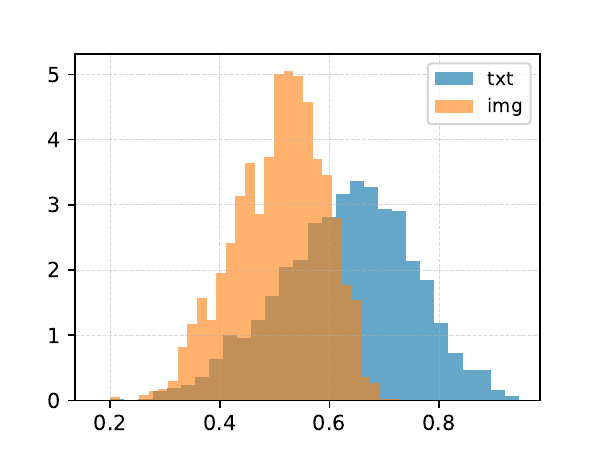}
      \caption{MMQA}
    \end{subfigure}
    \begin{subfigure}[b]{0.48\textwidth}
      \includegraphics[width=\linewidth]{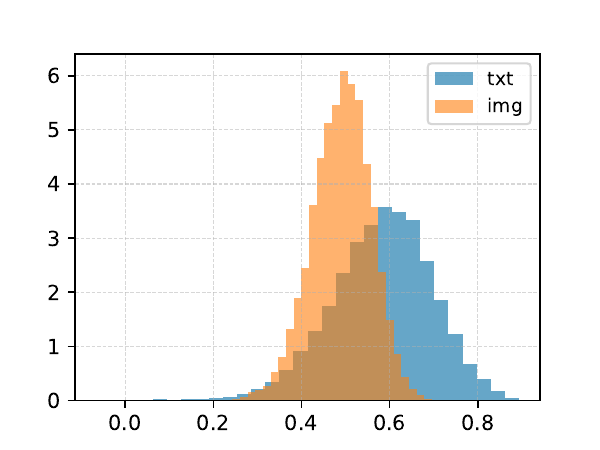}
      \caption{WebQA}
    \end{subfigure}
    \caption*{(G) Cohere Embed 3 English}
  \end{minipage}
  \caption{Distributions of cosine similarity scores between textual queries in the training split of each dataset and their corresponding examples~(either text or image).}
  \label{fig: cos_hists}
\end{figure*}

Figure~\ref{fig: cos_hists} presents the distributions of cosine similarity scores between textual queries~(from the training sets of MMQA and WebQA) and their positive examples, separated by the modality of positive examples. 

\clearpage

\end{document}